\documentclass[runningheads]{llncs}
\usepackage{graphicx}
\usepackage{amsmath,amssymb} % define this before the line numbering.
\usepackage{color}

\usepackage{bm}
\usepackage{color}
\usepackage{subfig}

\usepackage{amssymb}
\usepackage{bm}
\usepackage{scalerel}

\usepackage{array,multirow,graphicx}
\usepackage{multicol}
\usepackage{float}
\usepackage{subfig}
\usepackage{array}
\usepackage{makecell}

\usepackage{algorithm}
\usepackage{algorithmicx}
\usepackage{algpseudocode}

\makeatletter
\renewcommand{\Function}[2]{%
  \csname ALG@cmd@\ALG@L @Function\endcsname{#1}{#2}%
  \def\jayden@currentfunction{#1}%
}
\newcommand{\funclabel}[1]{%
  \@bsphack
  \protected@write\@auxout{}{%
    \string\newlabel{#1}{{\jayden@currentfunction}{\thepage}}%
  }%
  \@esphack
}
\makeatother

\usepackage{pbox}
\usepackage{wrapfig}
\usepackage{textcomp}

\makeatletter
\let\NAT@parse\undefined
\makeatother
\usepackage{url}
\usepackage{cite}

\makeatletter
\newcommand{\printfnsymbol}[1]{%
  \textsuperscript{\@fnsymbol{#1}}%
}

\makeatother

\begin{document}
\pagestyle{headings}
\mainmatter

\def\ACCV20SubNumber{839}  % Insert your submission number here

%===========================================================
\title{Mapping of Sparse 3D Data using \\ Alternating Projection} % Replace with your title
\titlerunning{Mapping of Sparse 3D Data using Alternating Projection}

\author{Siddhant Ranade$\thanks{indicates equal contribution.}$\inst{1}\and
 Xin Yu\printfnsymbol{1}\inst{1}\and
 Shantnu Kakkar\inst{2}\and \newline
 Pedro Miraldo\inst{3}\and
 Srikumar Ramalingam\thanks{Corresponding author.}\inst{1}} 
 \authorrunning{S. Ranade et al.}

 \institute {University of Utah, Salt Lake City UT 84112, USA  
\email{\{sidra,xiny,srikumar\}@cs.utah.edu} \and
 Trimble, Sunnyvale CA 94085, USA
\email{shantnu\_kakkar@trimble.com}\and
%\foreignlanguage{portuguese}{Instituto Superior Técnico}, 1049-001 Lisboa, Portugal 
Instituto Superior T\'ecnico, University of Lisboa, Portugal
 \email{pedro.miraldo@tecnico.ulisboa.pt}}

\maketitle

%===========================================================
\begin{abstract}
We propose a novel technique to register sparse 3D scans in the absence of texture. While existing methods such as KinectFusion or Iterative Closest Points (ICP) heavily rely on dense point clouds, this task is particularly challenging under sparse conditions without RGB data. Sparse texture-less data does not come with high-quality boundary signal, and this prohibits the use of correspondences from corners, junctions, or boundary lines.
Moreover, in the case of sparse data, it is incorrect to assume that the same point will be captured in two consecutive scans. We take a different approach and first re-parameterize the point-cloud using a large number of line segments. In this re-parameterized data, there exists a large number of line intersection (and not correspondence) constraints that allow us to solve the registration task. We propose the use of a two-step alternating projection algorithm by formulating the registration as the simultaneous satisfaction of intersection and rigidity constraints. The proposed approach outperforms other top-scoring algorithms on both Kinect and LiDAR datasets. In Kinect, we can use 100X downsampled sparse data and still outperform competing methods operating on full-resolution data.  
\keywords{LiDAR, 3D registration, line intersection, generalized relative pose estimation}
\end{abstract}

{
\setlength{\tabcolsep}{3.5pt}
\begin{figure}[t]
    \begin{center}
        \begin{tabular}[t]{ccc}
        \subfloat[\it Original point clouds from two 3D scans, down-sampled by 100x]{
        \begin{tabular}[t]{c}
        \includegraphics[width=.27\linewidth]{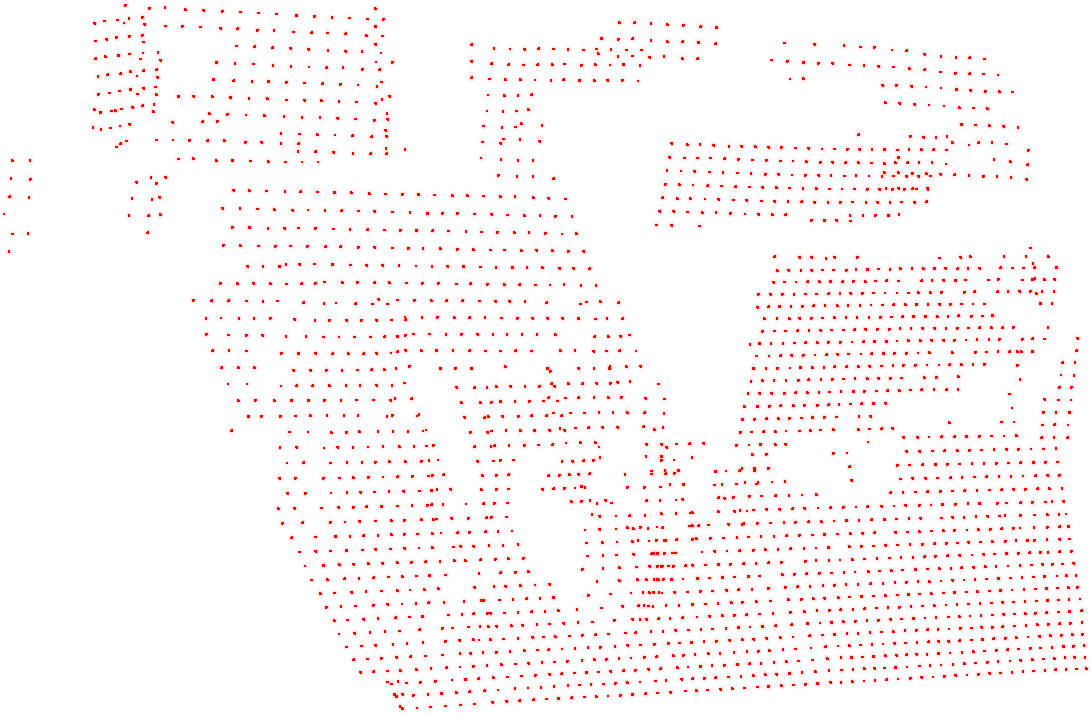} \\
        \includegraphics[width=.27\linewidth]{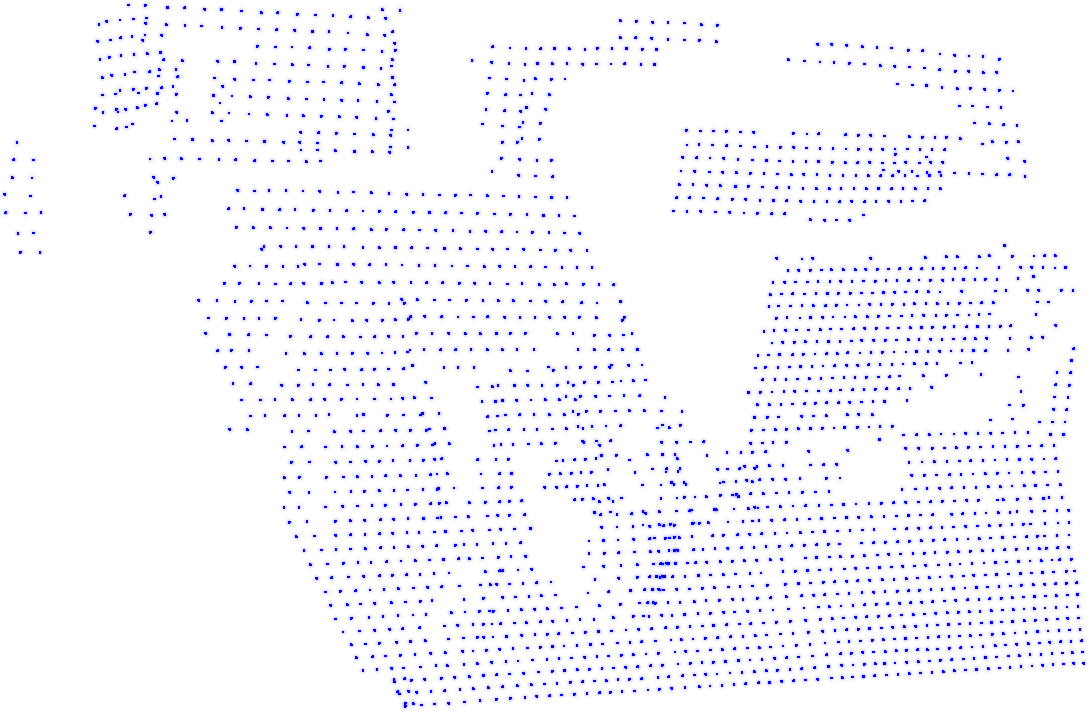}
        \end{tabular}
        \label{fig:introa}
        } &
        \subfloat[\it 3D straight lines fitted from the down-sampled frames \protect\subref{fig:introa}.]{
        \begin{tabular}[t]{c}
        \includegraphics[width=.27\linewidth]{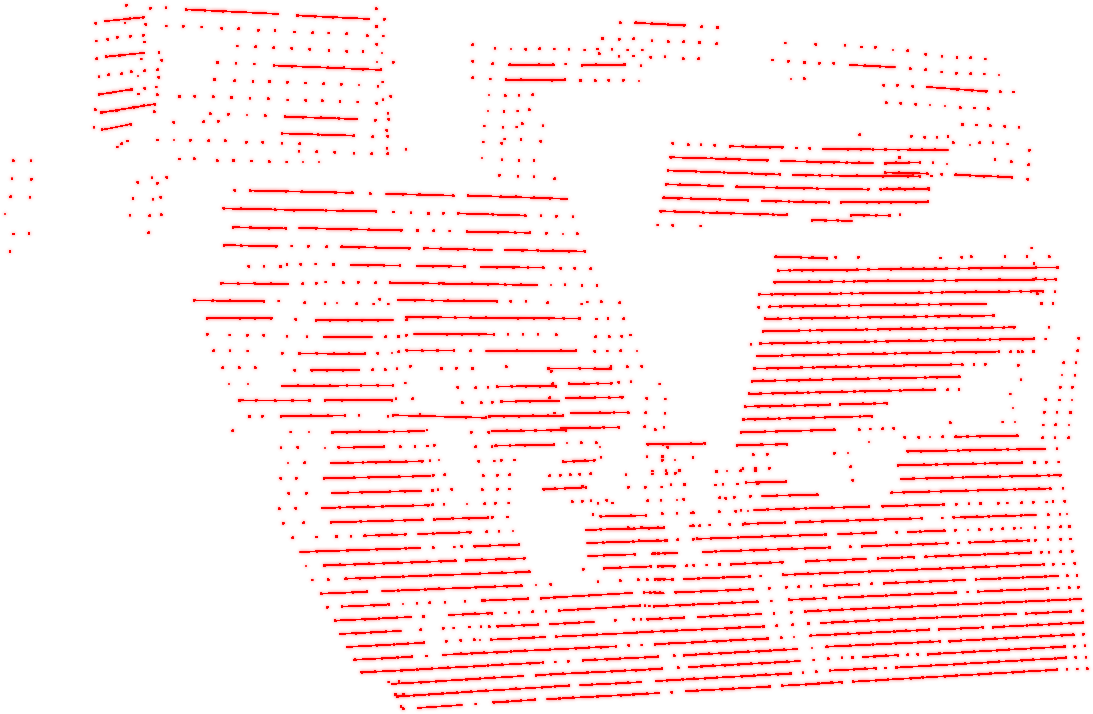} \\
        \includegraphics[width=.27\linewidth]{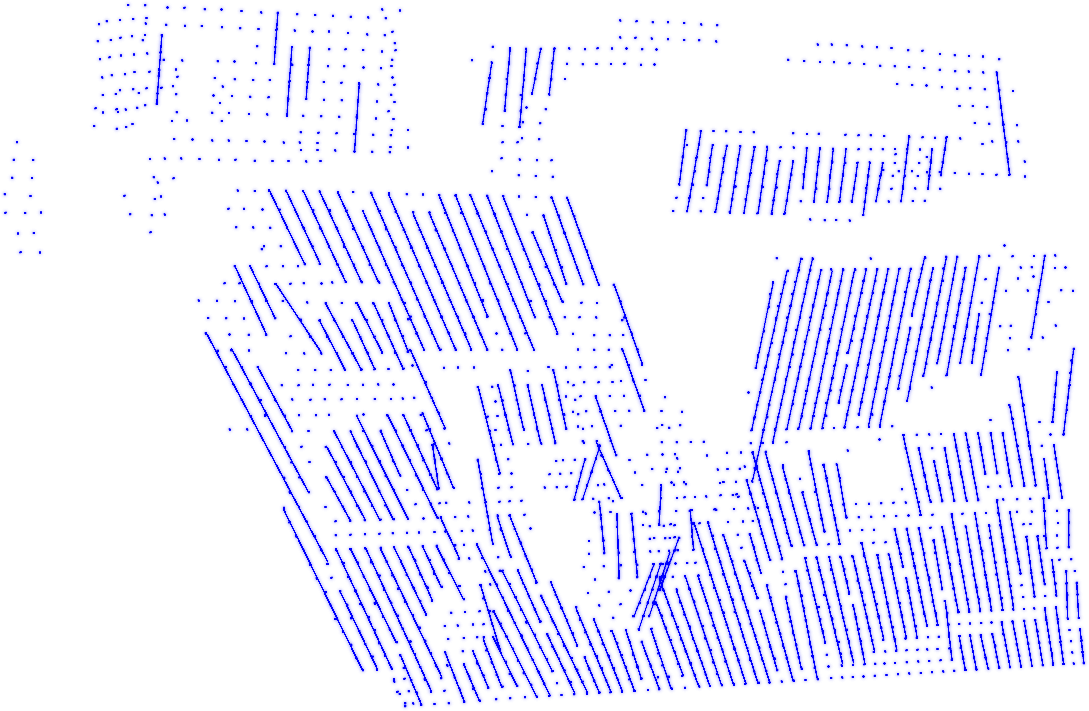}
        \end{tabular}
        \label{fig:introb}
        } &
        \begin{tabular}[t]{c}
        \subfloat[\it Aligned lines in \protect\subref{fig:introb}.]{
        \includegraphics[width=.27\linewidth]{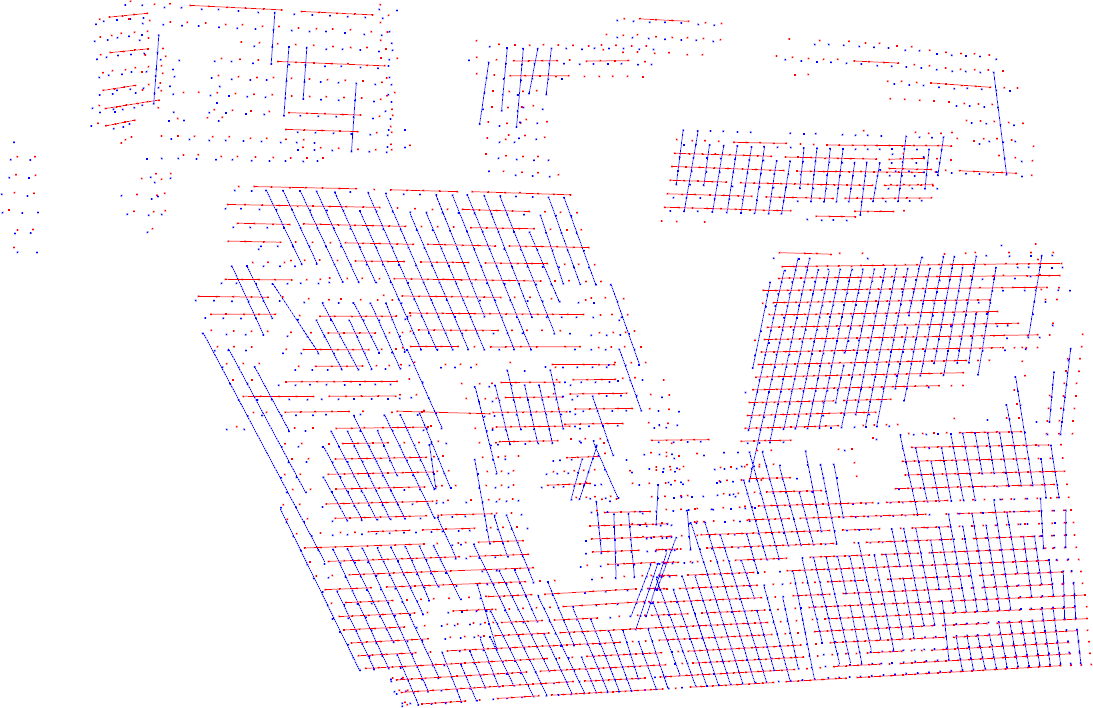}
        \label{fig:introc}
        }  \\
        \subfloat[\it Aligned point-cloud.]{
        \includegraphics[width=.27\linewidth]{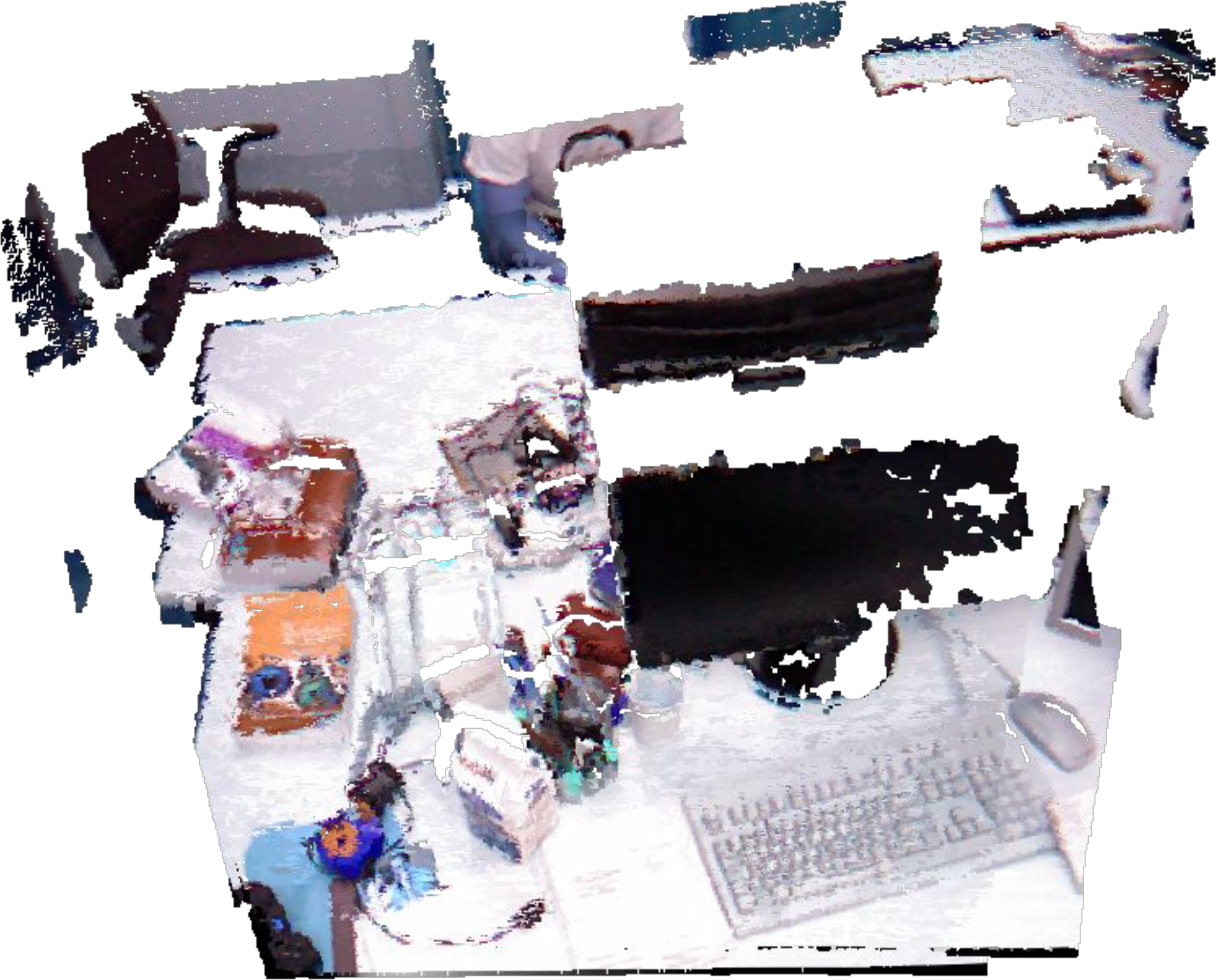}
        \label{fig:introd}
        } 
        \end{tabular}
        \end{tabular}
        \vspace{-.5cm}
    \end{center}
    \caption{\it Illustration of the proposed method. \protect\subref{fig:introa} shows 100$\times$ down-sampled Kinect depth data (64$\times$48). We fit 3D line segments to rows in the first and columns in the second frames as shown in \protect\subref{fig:introc}. We identify line-to-line intersection correspondences and use alternating projection~\cite{Bauschke96onprojection} to get the 3D scan alignment  \protect\subref{fig:introc}. \protect\subref{fig:introd} shows the registered point-cloud in original resolution with color. The average error is 0.53 degrees for rotations, and 7.58 mm for the translation, an improvement of 1.82x for the translation and 1.35x for the rotation, compared to FGR ~\cite{zhou16} which uses the full resolution Kinect data (640$\times$480) for registration.} 
    \label{fig:intro_figure}
\end{figure}
}

%%%%%%%%% BODY TEXT
\section{Introduction}

The last few years have witnessed the rise of inexpensive 3D sensors for both indoor (e.g., Microsoft Kinect) and outdoor scenes (e.g., Velodyne's VLP-16 LITE LIDAR, ZED stereo camera). The algebraic centerpiece of all 3D problems is the 3D point cloud registration. This problem is well studied in the case of dense settings and scenarios with point correspondences extracted from RGB components. Solutions include recent deep learning machinery~\cite{Pais2019}, the celebrated iterative closest point (ICP)~\cite{besl92}, and half a century old orthogonal Procrustes method~\cite{schonemann66} that is universally used till date. In this work, we study the problem of aligning highly sparse 3D scans without RGB information. In this setting, we lack boundary information, and without RGB components, it is hard to obtain any form of correspondences based on corners or line segments for registration.

We show the basic idea of our technique in Fig.~\ref{fig:intro_figure}. We fit lines to the data and represent the point-cloud using a large number of line segments. The original data hardly contains any correspondence constraints. On the contrary, in the re-parameterized line-based representation, there exists a rich set of line intersection constraints. Our strategy is to exploit these constraints to solve the registration problem. We consider 100x down-sampled Kinect data without RGB components. We show that the proposed algorithm can outperform competing Kinect registration algorithms that use full resolution (see Tab.~\ref{tab:TUM}). 

There is an interesting connection between our approach and the relative pose estimation for generalized cameras. The problem of relative pose estimation involves finding rotation and translation between two cameras such that the projection rays associated with corresponding 2D points intersect with each other. In the case of perspective cameras, we can find the motion up to a scale, and the associated projection rays are central~\cite{nister03}. In the case of generalized cameras~\cite{grossberg01}, the projection rays are unconstrained, and we are looking at the alignment of two sets of line segments such that the corresponding line segments intersect with each other. In our problem, we compute line segments from the point cloud representing the 3D scene. In both settings, we try to satisfy line intersection constraints (which essentially amounts to placing four 3D points on a plane), where the four 3D points are the endpoints of the intersecting line segments.  While minimal solvers typically produce robust solutions in general, the 6-point minimal solver for generalized relative pose \cite{Stewenius2005} has degeneracy problems (i.e., the constraints are sometimes insufficient to produce a unique solution, and thereby lead to infinite number of poses). The algorithm is supposed to produce unique pose for truly unconstrained set of projection rays (two or more projection rays from the same camera can not be parallel or coplanar). It is hard to expect that the line segments re-parameterizing the point-cloud will satisfy them. Thereby, we solve the registration using an alternating projection~\cite{Bauschke96onprojection} method ({\sc AP}) with seven pairs of line intersection constraints (one additional constraint is used since six intersection constraints can have up to 64 different solutions~\cite{Stewenius2005}).

In addition to line intersection constraints, we also utilize constraints that corner points, occurring in the boundary regions, are incident with edge lines in the scene. The highlights of the paper are the following:

\begin{enumerate}
\item We propose an {\sc AP}~\cite{Bauschke96onprojection} algorithm for the problem of scan alignment by showing that it is equivalent to the generalized relative pose estimation, i.e., pose estimation by satisfying intersection constraints among six or more pairs of corresponding line segments.
\item We show a family of registration algorithms that, additionally, exploit corner points in sparse point clouds and utilize collinearity (corner points lying on corner lines) constraints. 
\item We outperform other registration algorithms such as fast global registration (FGR) and Super4PCS, despite using 100X down-sampled point-cloud and not relying on the color information.
\item In 6 out of 11 KITTI LiDAR sequences, we outperform LOAM~\cite{Zhang14}, a competing LiDAR registration algorithm that minimizes point-to-line and point-to-plane distances in a Levenberg Marquardt framework for registration.
\item On down-sampled KITTI data, we outperform LOAM in all 11 sequences.
\end{enumerate}
    
\section{Related Work}

3D scan alignment is a classical problem in the computer vision and robotics communities, and there is a rich body of literature on this topic.

\vspace{.25cm}
\noindent

{\bf Iterative techniques:}
Iterative Closest Point (ICP) and its variants \cite{arun87,horn87,umeyama91,besl92,penney01,ROB-035} are the most applied scan alignment algorithm, and it works well for dense point-clouds with good initialization. ICP~\cite{besl92}, the gold standard for dense point-to-point registration, is an alternating minimization algorithm, alternating between finding closest correspondences and computing the rigid transformation. Alternating minimization and projection algorithms~\cite{Bauschke96onprojection} have been used for human pose and other geometrical problems~\cite{Zhou2015a,Zhou2015b,yan15,Schops2019,Campos2019}.

Techniques for finding globally optimal solutions combine local or probabilistic methods with graph optimization \cite{THEILER2015126} and branch-and-bound \cite{Campbell16,li17}. A closely related work is LOAM~\cite{Zhang14,Zhang17}, which is a top-ranking LiDAR alignment algorithm. We differ with LOAM as follows: 1) LOAM uses a distortion correction step for addressing the motion of the sensor at low frame-rate. This effect is similar to the rolling shutter effect in cameras. 2) LOAM uses the Levenberg Marquardt method for non-minimal registration, while we develop {\sc AP} algorithm for near-minimal configurations and utilize the RANSAC framework. 3) LOAM additionally utilize IMU, and we rely only on sparse 3D points.

Several global methods have been proposed in the literature, such as~\cite{yang13,yang16}. One of the problems with these methods is the high computation requirement for doing the branch and optimization. To overcome this, one line of research attempts to decompose the task of finding the relative transformation into first finding the rotation and then obtaining the translation given the optimal rotation \cite{Makadia06_cvpr,Straub17_cvpr}. Further, \cite{Liu18_eccv} proposes Rotation Invariant Features (RIF) to ease the task of decoupling rotation solution from the translation. However, none of the methods have been tested on LiDAR data.

\vspace{.25cm}
\noindent
{\bf Minimal Solvers:}
Other popular scan alignment methods in the absence of good initialization include minimal 3-point solvers in a RANSAC framework. The most common approaches to remove outliers from the data are based on the use of RANSAC~\cite{fischler81} plus some 3D point correspondences solver, such as \cite{schonemann66,Miraldo2019}. In addition to points, several registration algorithms have utilized other features on beam-based environment modeling~\cite{endres14}, 3D planes~\cite{Raposo13plane-basedodometry,Zhou-2018-107715,ma16,bhattacharya17,liu18,Grant2018}, 3D line segments~\cite{lu15,Zhou-2018-107715}, implicit surface representation~\cite{Deschaud18}, and edges~\cite{choi13}. A detailed survey on 3D SLAM methods can be found in ~\cite{endres12,sturm12}.

Our approach for the registration of 3D scans is tightly connected with minimal solvers for relative pose estimation. In particular, we have relative pose estimation algorithms for calibrated perspective cameras~\cite{nister03,li06b}, with known relative rotation angle~\cite{li13}, with known directions~\cite{fraundorfer10,saurer15}, with unknown focal lengths~\cite{stewenius05b,li06}, solutions invariant to translation~\cite{kneip12}, and generalized relative pose~\cite{Stewenius2005,ventura15}. Recently, a minimal hybrid solver considers relative and absolute poses~\cite{camposeco18}. In the case of absolute poses, we have perspective ~\cite{kneip11,ke17,wang18,persson18}, and multi-perspective systems~\cite{ventura14,camposeco16}. For 3D scans, we have the Procrustes solver for a pairwise alignement~\cite{schonemann66}, and solutions to the mini-loop alignments in~\cite{Miraldo2019}. It was recently shown that the generalized relative pose estimation~\cite{pless03,Stewenius2005,sturm05,Li2008,Kneip2014} can be used in privacy-preserving pose estimation from 3D pointclouds~\cite{Pittaluga_2019_CVPR}.

\vspace{.25cm}
\noindent
{\bf Deep scan alignment:}
Deep neural networks (DNNs) for solving 3D registration problems have increased significantly in the last few years. Several methods have been developed to extract local 3D geometric structures, such as \cite{elbaz17,khoury17,Zhou2018b,Deng2018,Deng2019,Lu_2019_ICCV}. There have been a few recent algorithms for LiDAR registration~\cite{Ding2019,Elbaz2017} using deep neural networks, but they are mostly applicable to dense point clouds.

A technique for registering 3D point clouds given a set of point correspondences is proposed in~\cite{Pais2019}. Floor plan reconstruction using deep networks has been shown in~\cite{liu18}. The PointNet \cite{Qi2017} is used in \cite{Aoki2019}, together with Lukas-Kanade algorithm in a single network. The proposed method works in an iterative fashion, similar to ICP. In ~\cite{Wang_2019_ICCV,Lu_2019_ICCV}, a differential SVD layer is coupled with DNNs for generating matching. DNNs also prove to be helpful for solving other kind of problems such as multi-scan transformation averaging~\cite{Chatterjee2018}. \cite{huang19} utilizes deep neural networks to compute the weights for different pairwise relative pose estimates and \cite{Ding2019} use two networks, for pose and scene estimation.
\section{Problem Statement}

Given two sparse 3D point sets $\{p_1^1,\dots,p_{n_1}^1\}$ and $\{p_1^2,\dots,p_{n_2}^2\}$, our goal is to compute $T = (R, \mathbf{t})$, such that $\{p_1^1,\dots,p_{n_1}^1\}$ and the transformed point set $\{Rp_1^2+\mathbf{t},\dots,Rp_{n_2}^2+\mathbf{t}\}$ would model or represent the same 3D scene in the first coordinate frame.

We model the scene using a set of 3D line segments on planar surfaces and the detection of a few informative 3D corner points, and 3D edge lines. Concretely, the first point set  $\{p_1^1,\dots,p_{n_1}^1\}$  is re-parameterized as $\{l_1^1,\dots,l_{m_l}^1,k_1^1,\dots,k_{c_1}^1\}$, where $l$ denotes \textit{lines} and $k$ are \textit{corners} (obtained by line fitting and corner estimation respectively, see Sec.~\ref{subsec:preprocessing}). The second point set $\{p_1^2,\dots,p_{n_2}^2\}$ is re-parameterized as  $\{g_1^2,\dots,g_{m_2}^2,q_1^2,\dots,q_{c_2}^2\}$, where $g$ denotes \textit{lines} and $q$ are \textit{edges} (obtained by line fitting and edge line estimation respectively, see Sec.~\ref{subsec:preprocessing}). In the original point set representation, we can not identify any point correspondences due to sparsity. The re-parameterization allows us to obtain a rich set of line intersections (e.g., lines $l_i^1$ and $g_i^2$ intersect) and incidence relations (e.g., an edge $q_j^2$ passes through a corner point $k_j^1$).

Our implementation considers line intersections, and corner-edge, edge-corner incidences. For clarity of illustration, we will only look at lines and corners from the first frame, and lines and edges from the second.

\begin{figure*}[h]
    \begin{center}
        \includegraphics[width=.9\linewidth]{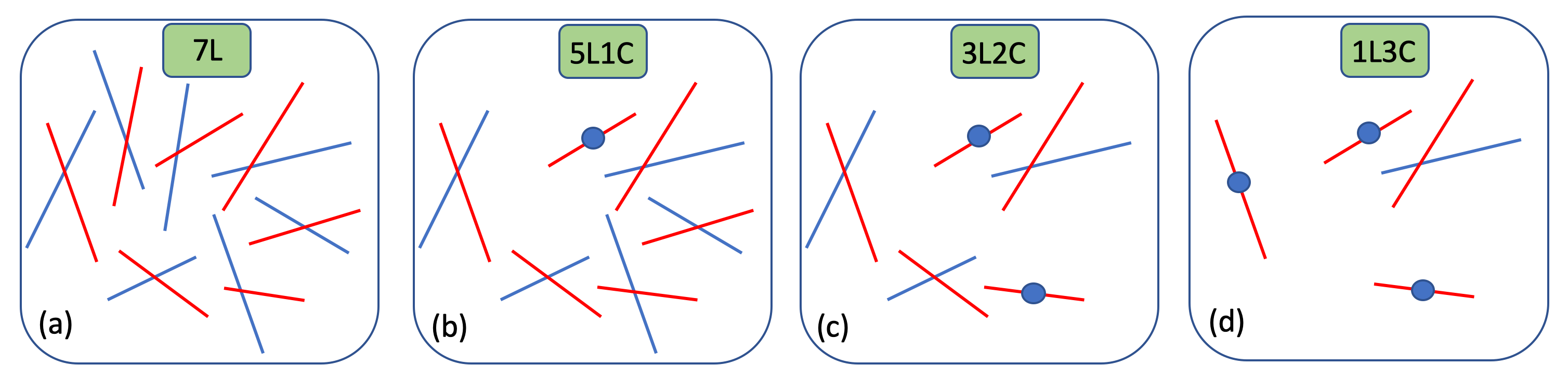}   % 
    \end{center}
    \caption{\it The four sets of constraints addressed in this paper are denoted by 7L, 5L1C, 3L2C, and 1L3C. For example, 3L2C in (c) denotes a total of five constraints (three line-to-line intersections, and two corner-to-line incidences). For simplicity, we consider the corners only from the first frame. The primitives from the first and the second frame are shown in blue and red, respectively.}
    \label{fig:four_solvers}
\end{figure*}

In Fig.~\ref{fig:four_solvers} we show the different sets of constraints used in the paper. The line-to-line intersection constraint enforces that the four end-points of two intersecting line segments are coplanar, i.e., lie on the same plane. The corner-to-line incidence constraint enforces that a corner point lies on a line. It is always a useful exercise to check the degrees of freedom ({\sc dof}) of the unknowns and the number of available constraints. Our goal is to compute 6 {\sc dof} pose $(R,\mathbf{t})$ using line-to-line intersections and point-to-line incidences. Each constraint of line-to-line intersection and point-to-line incidences takes away 1 {\sc dof} (See ~\cite{Stewenius2005}) and 2 {\sc dof}s (See ~\cite{nister03,li06b}), respectively. The algebraic methods that employ exact minimal solvers (where the {\sc dof}s from the constraints and the unknowns are equal) typically produce multiple solutions for poses corresponding to the multiple roots of the higher degree polynomial systems. Consequently, it needs a second step with additional correspondence constraints to pick the correct pose from the multiple solutions. In contrast to the algebraic solvers, we use the {\sc AP} algorithm on near-minimal cases (i.e. each of the 4 sets of constraints shown in Fig.~\ref{fig:four_solvers} has one {\sc dof} extra over the six unknowns), and the over-constrained setting leads to a single pose.

\section{Alternating Projection ({\sc AP})}

\begin{figure*}[h]
    \begin{center}
        \includegraphics[width=.9\linewidth]{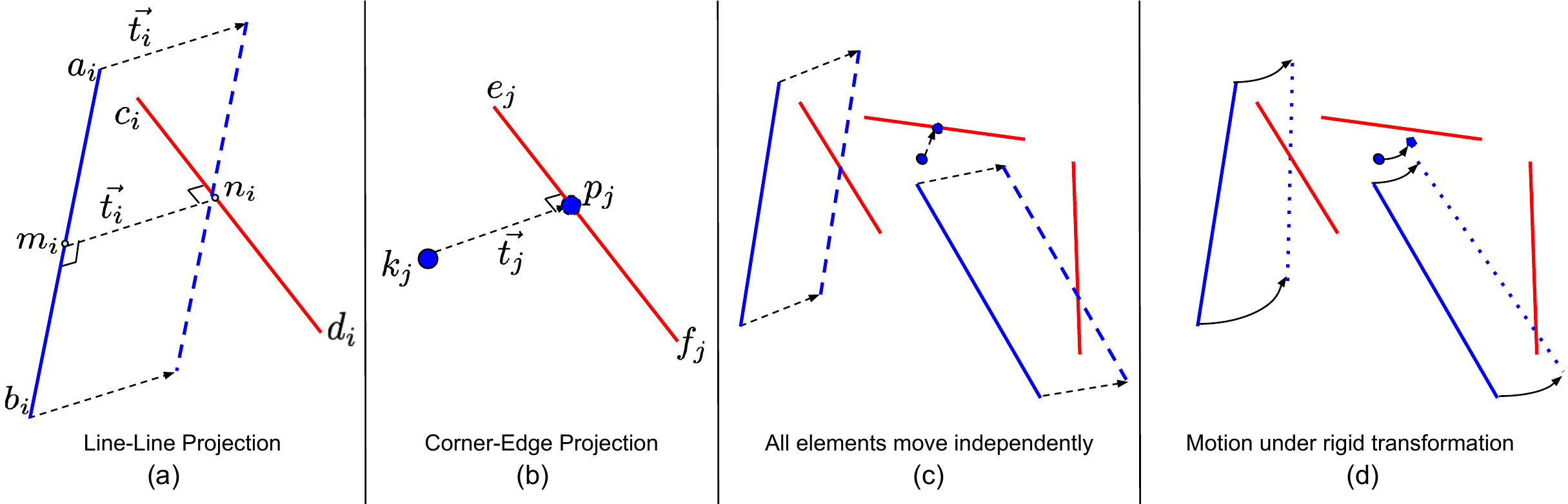}   % https://docs.google.com/drawings/d/1cIVYmq37LMnTq_aoe6nGPRHeRnK4R3_EdoEWG-E_T3g/edit?usp=sharing
    \end{center}
    \caption{\it 
    %\textcolor{red}{ The simplest way to satisfy their intersection constraints, for all input pairs (both line-line and corner-edge) is to translate elements (lines, corners) from the first frame towards their counterparts (lines, edges) in the second frame, independent of the other pairs, as illustrated in (a), (b), and (c).  However, after applying this projection operator for enforcing the intersection constraint, we  does not preserve the rigidity constraint for the first frame as a whole. Thus the projection operator for rigidity constraint would first computes a single rigid transformation that the lines and corners from the first frame as close as possible to their projected positions using Orthogonal Procrustes, as shown in (d). We repeat these two steps (intersections followed by rigid transform estimation) until the update is lower than a threshold.}
    %\textcolor{blue}{
    One {\sc AP} iteration on a set of line-line and corer-edge pairs has two steps: 1) To satisfy their {\textit{\bf intersection}} constraints, it translates elements (lines, corners) from the first frame towards their counterparts (lines, edges) in the second frame, independent of the other pairs, as illustrated in (a), (b), and (c). % However, after applying this projection operator for enforcing the intersection constraint, we does not preserve the rigidity constraint for the first frame as a whole. 
    2) To preserve the {\bf\textit rigidity} for the first frame elements as a whole, it computes a single rigid transformation that takes all elements in this frame as close as possible to their position after the independent translation, as shown in (d). %We repeat these two steps (intersections followed by rigid transform estimation) until the update is lower than a threshold.
    %}
    }
    \label{fig:ap_vis}
\end{figure*}

We briefly explain the general approach used in {\sc AP}~\cite{Bauschke96onprojection}. We compute a  point (a higher dimensional vector denoting the unknowns) in the intersection of sets (denote the various constraint sets) by a sequence of projections onto the sets. Let us denote our unknown parameter vector as $x \in \mathbb{R}^n$, and the two sets as $C$ and $D$ in $\mathbb{R}^n$, respectively. Let $P_C$ and $P_D$ be projection operators. Given a point $x \in \mathbb{R}^n$, the projection operator $P_C(x)$ produces a point $x' \in C$ such that $|x-x'|$ is minimum. The key elements of the {\sc AP} are the constraints and the projection operations. 

In our registration setting, given a set of candidate line intersections and corner-edge incidences, the goal of {\sc AP} is to estimate a rigid transformation which satisfies the given intersection/incidence constraints (henceforth referred to simply as intersection constraints). Thus the solver needs to compute the unknowns $(R,\mathbf{t})$ to satisfy two constraint sets: {\bf intersection}, and {\bf rigidity}. The projection operators for intersection and rigidity constraints are shown in Fig.~\ref{fig:ap_vis}.

\begin{algorithm}
\caption{Alternating Projection}
\label{alg:AM}
\begin{algorithmic}[1]
\Require {$S^1=\{l_1^1,..,l_M^1, k_1^1,..,k_N^1\}$,
$S^2 = \{g_1^2,..,g_M^2, q_1^2,..,q_N^2 \}$}
\Ensure{$T$}
%\State $\delta = \infty$
\Repeat
%\While{$\delta > \epsilon$} \Comment{$\epsilon$ is a distance threshold}
    \State $\overrightarrow{t_i}$ = LineLineProjection($l_i^1, g_i^2$) for i $\in$ $1,...,M$ \Comment{as shown in ~Fig. 3(a)}
    \State $\overrightarrow{t_j}$ = CornerEdgeProjection($k_j^1,q_j^2$) for j $\in$ $1,...,N$ \Comment{as shown in ~Fig. 3(b)}
    \State $l_i^{\prime1} = l_i^1 + \overrightarrow{t_i}$ for i $\in$ $1,...,M$ \Comment{as shown in ~Fig. 3(c)}
    \State $k_j^{\prime1} = k_j^1 + \overrightarrow{t_j}$ for j $\in$ $1,...,N$
    \State $S^{\prime1} = \{l_1^{\prime1},..,l_M^{\prime1}, k_1^{\prime1},..,k_N^{\prime1}\}$
    \State $T = $\small{\texttt{RigidAlign}}($S^1, S^{\prime1}$) 
    \State Update $l_i^1 = T \cdot l_i^1$ for i $\in$ {1,...,M}
    \Comment{as shown in ~Fig. 3(d)}
    \State Update $k_j^1 = T \cdot k_j^1$ for j $\in$ {1,...,N}
    %\State $ LD_i = dist(l_i^1, l_i^2) $ for i $\in$ $1,...,M$
    %\State $ PD_j = dist(k_j^1, q_j^2) $ for j $\in$ {1,...,N}
    \State $\delta = \max\limits_{i,j}(dist(l_i^1, g_i^2), dist(k_j^1, q_j^2))$
%\EndWhile. \Comment{Or if max iterations reached}
\Until {$\delta \le \epsilon$} or max iterations reached \Comment{$\epsilon$ is a distance threshold}
\State \Return $T$
\end{algorithmic}
\end{algorithm}

Formally, let us denote the lines from the first frame with $l_i^1 = (a_i, b_i)$ (where $a_i$ and $b_i$ are the end-points), and those from the second frame by $g_i^2 = (c_i, d_i)$, for $i \in \{1, \ldots, M\}$, i.e. each $(l_i^1, g_i^2)$ is a candidate pair of lines, where $M$ is the number of line intersections used by that particular solver. For simplicity, let's consider only corners from the first frame $k_j^1$ and edges $q_j^2 = (e_j, f_j)$ from the second, for $j \in \{1, \ldots N\}$, such that each $(k_j^1, q_i^2)$ is a candidate corner-edge pair, where $N$ is the number of corner-edge incidence constraints used by the solver. In this paper, we solve the cases $(M, N) \in \{(7,0), (5,1), (3,2), (1,3)\}$, corresponding to our solvers 7L, 5L1C, 3L2C, and 1L3C respectively. Please note that edges are also line segments, but they are extracted in a different manner compared to regular line segments (see Sec.~\ref{sec.linefitting_matches}).

We illustrate our {\sc AP} solver in Fig.~\ref{fig:ap_vis} with a simple scenario that considers only two intersection constraints (one line-to-line and one corner-to-edge). In \ref{fig:ap_vis}(a), the closest points on the line segments are denoted by $m_i$ and $n_i$ respectively. The line segment from the first frame (\textcolor{blue}{blue}) is moved by the smallest distance so that the two line segments intersect. In \ref{fig:ap_vis}(b), $p_j$ denotes the point on the edge-line $q_j$ closest to the corner point $k_j$, i.e. $p_j$ is the projection of point $k_j$ on the line $q_j$. 
%\textcolor{red}{We first applying this projection operation to frame 1 (\textcolor{blue}{blue}) in (c),and then compute a single rigid transformation that takes the lines and corners from the first frame as close as possible to their projected positions using Orthogonal Procrustes in \ref{fig:ap_vis}(d).}
%\textcolor{blue}{
After applying this projection operation to frame 1 (\textcolor{blue}{blue}) in (c), we get their projected position. 
For the rigid alignment between frame 1 (lines, corners) and their projected positions, we estimate translation by aligning their centroids and then estimate rotation using Orthogonal Procrustes, as shown in \ref{fig:ap_vis}(d). We repeat these two steps (intersections followed by rigid transform estimation) until the update is lower than a threshold.
%}
%and then compute a single rigid transformation that takes the lines and corners from the first frame as close as possible to their projected positions using Orthogonal Procrustes in \ref{fig:ap_vis}(d). 
The complete algorithm is illustrated in Algorithm~\ref{alg:AM}. In our implementation, we move both elements of each pair towards each other, rather than moving only elements from the first frame towards those from the second.

\section{Implementation Details}
% \section{Line fitting and correspondences}
\label{sec.linefitting_matches}

\subsection{Pre-processing}
\label{subsec:preprocessing}

\noindent
\setlength{\intextsep}{2pt}%
\setlength{\columnsep}{7pt}%
\begin{wrapfigure}{r}{0.4\textwidth}
    \vspace{-2cm}
% \begin{figure*}[htb]
    \centering
    \subfloat[original scan]{\includegraphics[width=1\linewidth]{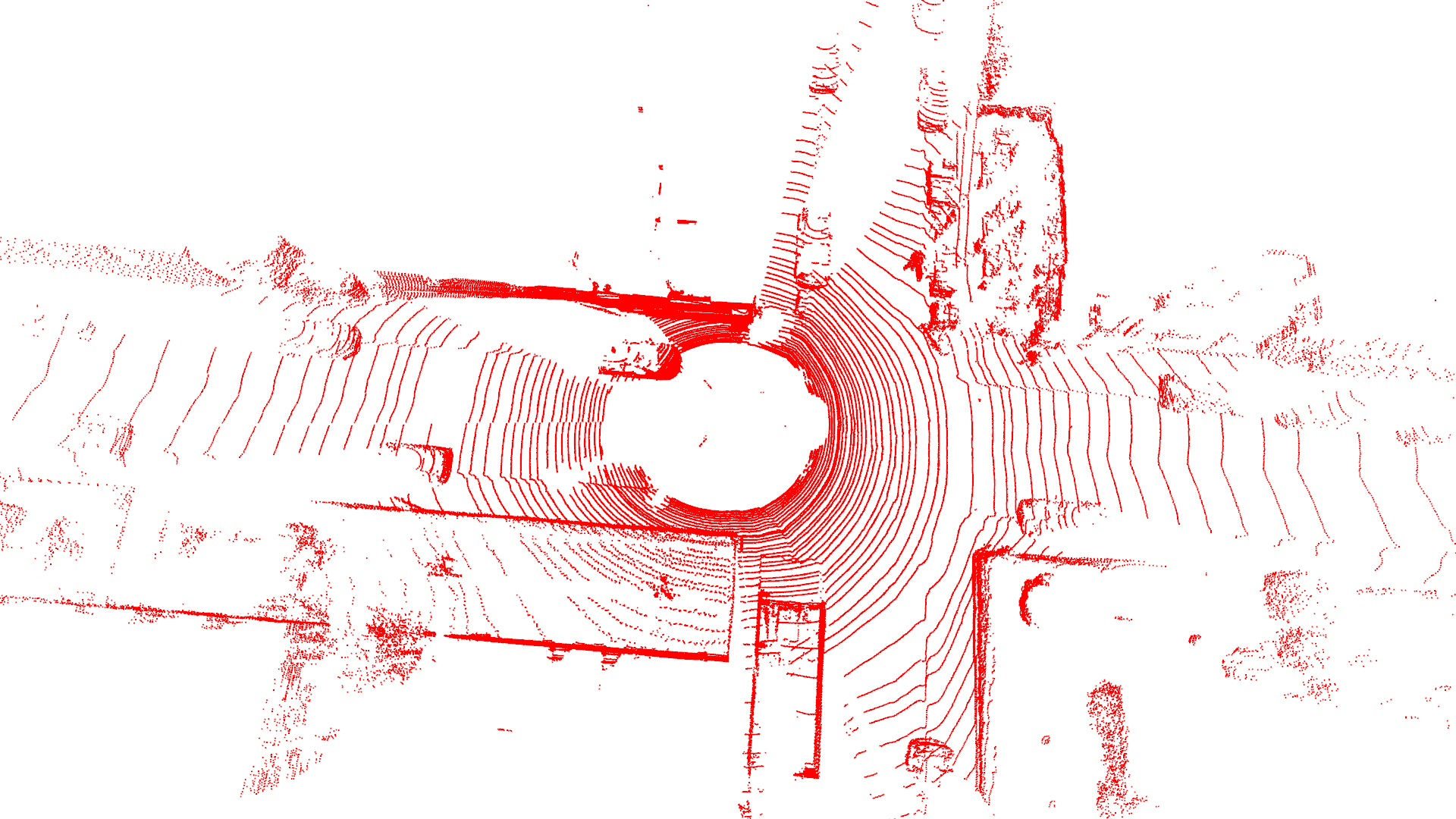}}\\
    \subfloat[1/36 downsampling]{\includegraphics[width=1\linewidth]{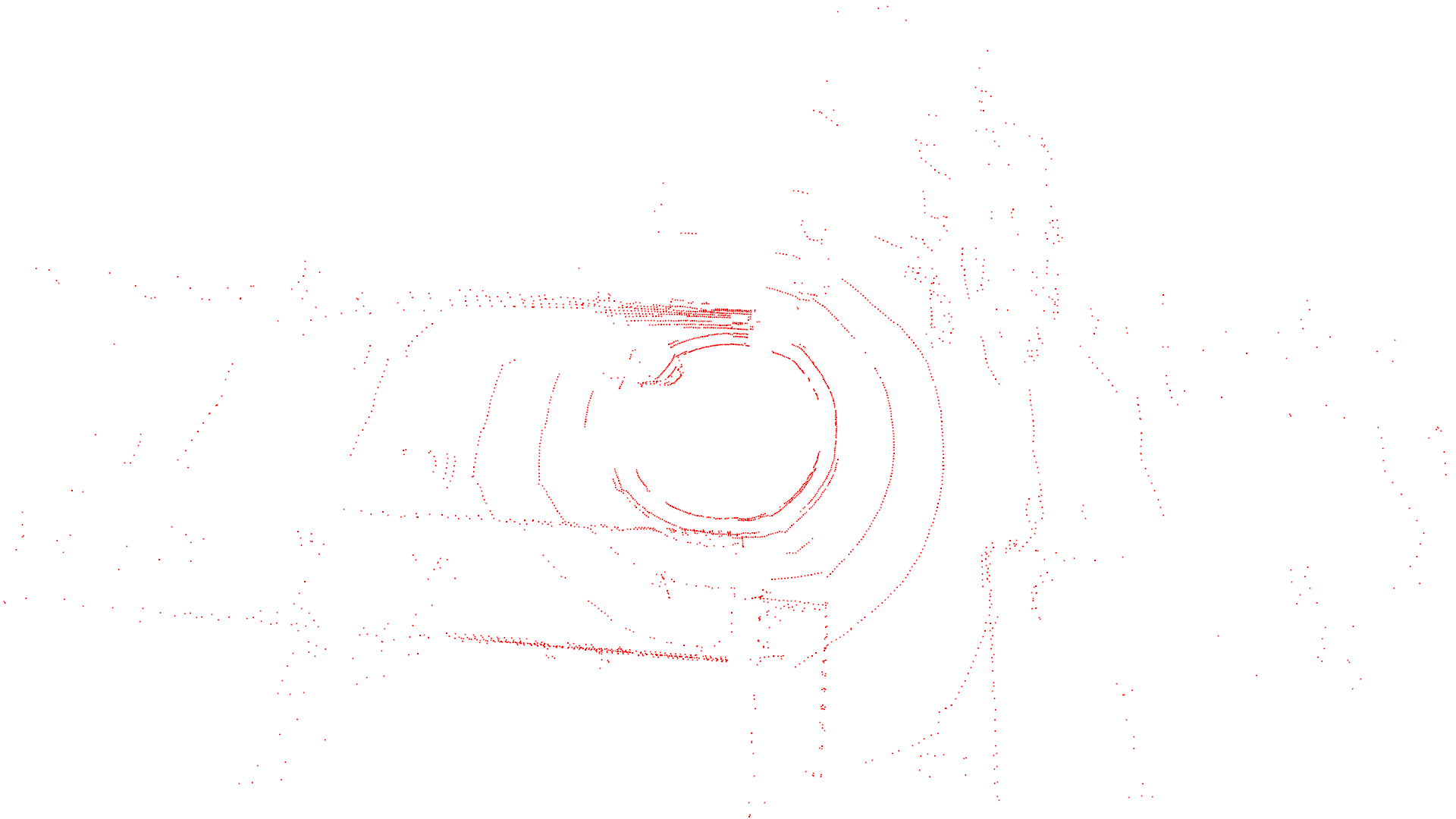}}
    \caption{\it Single frames from KITTI at (a) original resolution and (b) 1/36 down-sampled.}
    \label{fig:downsampling}
% \end{figure*}
\end{wrapfigure}
\setlength{\intextsep}{0pt}%
\setlength{\columnsep}{7pt}%
\begin{wrapfigure}{r}{0.4\textwidth}
% \begin{figure}[ht]
\vspace{-.25cm}
    \centering
    \includegraphics[width=\linewidth]{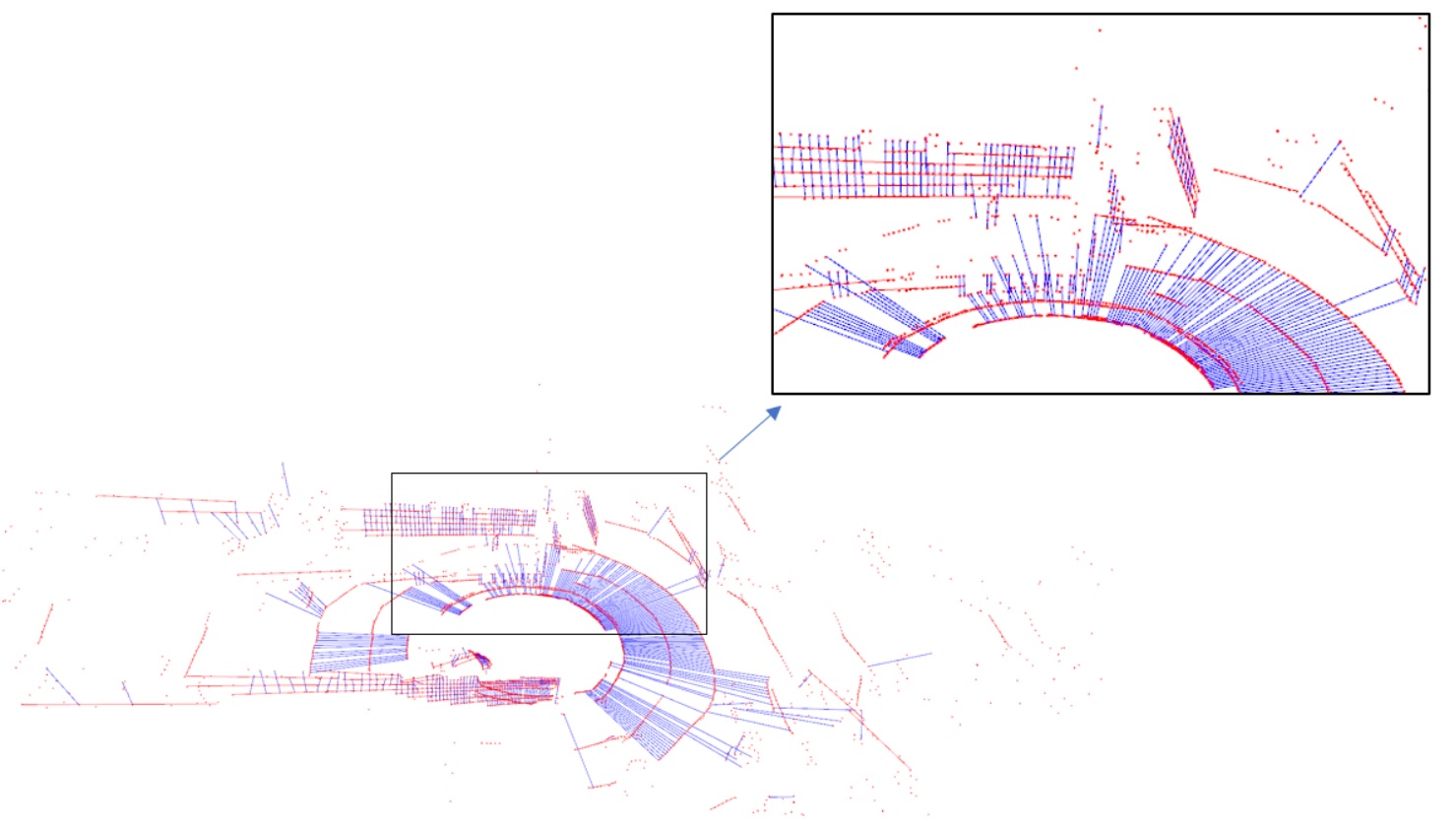}
   % \caption{\it Line fitting on one KITTI frame. Red lines come from horizontal scan-lines and blue from vertical. We call them H-lines and V-lines respectively.}
   \caption{\it Line fitting on one KITTI frame. Red lines come from horizontal scan-lines(\textcolor{red}{H-lines}) and blue from vertical(\textcolor{blue}{V-lines}).}
    \label{fig:line-fiting}
% \end{figure}
\end{wrapfigure}
\textbf{Organized point clouds:} Our algorithms require line 
fitting outputs, which we find is easiest done on organized point clouds. In the case of Kinect data, the input is already organized in the form of a depth image, and no further action is necessary. In the case of the KITTI data, we use the sensor calibration parameters to organize the raw input points into a grid-like structure by azimuth and elevation. In the experiments involving down-sampling of this data, we select points at appropriate indices from these organized point clouds. For instance, Fig.~\ref{fig:downsampling} shows a point-cloud down-sampled by a factor of 6 along both horizontal and vertical directions, to retain roughly 1/36\textsuperscript{th} the points.

\vspace{.25cm}
\noindent
\textbf{Line Fitting:} Given a set of 3D points in a regular grid (with some points missing due to sensor noise), we consider horizontal and vertical scan-lines in this organized point cloud and use RANSAC to fit lines to every scan-line. We call lines that come from horizontal scan-lines ``H-lines'', and the ones coming from vertical scan lines are called ``V-lines''.
% \begin{wrapfigure}{r}{0.5\textwidth}

For instance, in Fig.~\ref{fig:line-fiting}, H-lines are represented by the color red, and V-lines by blue. H and V do not refer to the orientations of the lines in 3D, only the scan-line they come from; the blue lines on the ground plane in this figure are actually V-lines. 
% Using intersections between these H- and V-lines (coming from the same frame), we can compute the normals for all lines, which we use in candidate selection step. All lines detected here come from some plane in the underlying scene, and these computed normal vectors are normals of those planes.

\vspace{.25cm}
\noindent
\textbf{Corner points and edge lines:} We use a similar formulation to~\cite{Zhang14} for estimating corner points. Let $\{x_i | i \in W\}$ denote the coordinates of the (ordered) points in a single scan-line in an organized point cloud in the local coordinate frame, where $W$ is the total number of points in the scan-line. We define a local smoothness term for each point $i$ based on $K$ neighbors on either side of $i$ as
\begin{equation}
    c_i = \frac{1}{2K \|x_i\|} \| \sum_{j=-K, j \neq 0}^K (x_i - x_j) \|,
\end{equation}
where a higher $c$ value indicates lower smoothness. We divide each scan-line into 4 zones, and select two points with the highest $c$ as corner points. We fit lines to corner points from successive 
%\textcolor{red}{scan lines} \textcolor{blue}{scans} 
scanes in the same frame to get the edge lines.

%% \vspace{.25cm}
%% \noindent
%% \textbf{Plane Fitting:} We use RANSAC for plane fitting as well. %For KITTI dataset, we apply 200 trials on each  frame  and  accept  one  plane  if  it  has  more  than  500 inlier points, which are within 8 cm from the plane.
%% KITTI has dense points on the ground. At least the ground plane will be detected, which is enough for the 3L1P method. %There can be more than one ground planes detected considering we don't remove rolling shift inside one frame. 
%% Figure~\ref{fig:plane_fitting} shows a good example of plane fitting, with sufficiently many planes in front and at the back and on the sides of the street.

% \begin{figure}[ht]
%     \begin{center}
%         \psfig{figure=./figs/plane_fitting.jpg,width=1.0\columnwidth}
%     \end{center}
%     \caption{Visualization of plane fitting on one KITTI frame. 10 planes are detected, with 8 above the ground.} 
%     \label{fig:plane_fitting}
% \end{figure}

\subsection{Full Meta-Algorithm with RANSAC}\label{sec:Meta-Algorithm}

At each step, we estimate the frame-to-frame transformations between successive frames. We designate the local frame of the first scan to be the world coordinate frame, and compose (multiply) these transformations to obtain transformation matrices for all frames with respect to the first frame.

% All three of our solvers, 7 line intersections (7L), 1 line intersection with 2 plane correspondences (1L2P), and 3 line intersections with 1 plane correspondence (3L1P) produce a transformation matrix given a small number of inputs, with the latter two being minimal solvers. Thus, we apply these solvers in a RANSAC framework to obtain the best possible transformation matrices. 
The proposed solvers (7L, 5L1C, 3L2C, 1L3C) produce a transformation matrix given a small number of inputs. We apply these solvers in a RANSAC framework to obtain the best possible transformation matrices. At each RANSAC iteration, we pick a solver at random (all $4$ solvers are selected with the same probability, i.e., $0.25$), and select candidates for that solver as appropriate.

\vspace{.25cm}
\noindent
\textbf{Candidates for RANSAC:} We identify candidates for (a) line intersection constraints based on the distance between line segments in the first and second scans, and (b) corner point-edge line incidence constraints based on the distances between corner points from one scan and edge lines from the other. We select from among these candidates at random in every RANSAC iteration. This simple approach assumes that the relative transformation is close to identity, but we observe that this works well in practice. 

% get these candidates based on using the current estimate of the relative transformation between successive frames (initially the identity matrix). {\color{blue}PM: should we add some papers in the literature that do the same? And possibily say that we could use other ideas?}

% \vspace{.25cm}
% \noindent
% \textbf{Candidate line intersections:} We consider all H-lines from one frame and V-lines from the other. Pairs of lines that are within a distance of some threshold (it is 2m for KITTI dataset) are considered to be candidate pairs.

% \vspace{.25cm}
% \noindent
% \textbf{Candidate plane correspondences:} Among all the possible plane pairs from two frames, the candidate pairs have the angle of the plane normals smaller than 20 degree, and have the distance from the centroid of one plane to another plane within the same threshold used for candidate line intersection. 

% \vspace{.25cm}
% \noindent
% \textbf{Candidate Selection:} It is necessary to select a good set of 7 line pairs in order to avoid degenerate cases. We cluster all the normal vectors from the first frame into 3 directions, and select 2 pairs uniformly at random from each cluster. The last pair is sampled (also uniform, random) from the full set of candidates, to get $2\times 3 + 1 = 7$ pairs.% Similarly, for algorithm 3L1P, we sample one pair from each cluster. Plane correspondences are sampled uniformly at random from the full set of candidates.

\vspace{.25cm}
\noindent
\textbf{Inlier counting:} % All three algorithms use line intersections to do inlier counting.
From the set of candidates pairs of lines, we count the number of pairs that have a distance less than another threshold (2 cm on KITTI dataset and 5mm on TUM here) after applying the computed transformation. %Lines whose normals belong to clusters with a smaller number of lines get a higher weight, and {\it vice versa}.

In addition, we find that we get better results by running the full algorithm described above thrice, each time using the best transformation from the previous step as the initial guess for candidate estimation. %\textcolor{red}{instead of the identity transform.} \textcolor{blue}{.} %(instead of the identity transform). 
With improved initialization we obtain better candidates for line intersection constraints. 
%\textcolor{red}{The use of better candidates lead to better set of inliers. In that sense, we can also treat this as some variant of ICP.} 
%\textcolor{blue}{
Both ICP and our algorithm iteratively update the relative transformation between two frames, but there are some important differences: (a) our algorithm needs only 3 iterations, while ICP typically needs many more, (b) within each iteration, our algorithm uses RANSAC, the AP solver  works with near-minimal sets of constraints, while ICP uses correspondence constraints involving all primitives.
%}

% In addition to the RANSAC, we find that in practice, we get better results by running the full algorithm described above thrice, each time using the best transformation from the previous step as the initial guess for candidate estimation. As the candidates improve, so do the inliers.

%%%%%%%%%%%%%%%%%%%%% 

% Our line intersection constraints are equivalent to co-planarity constraints on the four end-points of the lines. Using 7 pairs of lines for determining a transformation leads to a unique solution as long as the 7 pairs of lines correspond to at least 3 non-parallel planes. As an example, if all 7 pairs come from planes parallel to each other, the solution can only be obtained up to two degrees of freedom, i.e. in-plane translation. To overcome this limitation, we also make use of corner points and edge-lines. Similar to~\cite{Zhang14,Zhang17}, we identify corner points on every scan-line based on local smoothness. We then extract edge-lines by fitting line segments to corner points from successive scan-lines in the same scan. Then, given corner points from one scan and edge-lines from the other, we identify candidate corner-edge incidences, and incorporate these in the Alternating Projection in addition to the line-intersection constraints.
% \input{files/exp.tex}           % experimental setup
\section{Experiments}

We evaluated the proposed algorithms on two datasets: TUM~\cite{sturm12} and KITTI~\cite{Geiger2012CVPR}.

\noindent
\textbf{Relative Pose Error (RPE):} As proposed in \cite{sturm12}, we compute the error in the relative pose between successive frames, and report the translation error in meters and rotation error in degrees respectively.

\vspace{.15cm}
\noindent
\textbf{Translation and Rotation Error along the trajectory:} This metric is used in \cite{Geiger2012CVPR} and on their online leaderboard. For all sub-sequences of length 100m, 200m, $\ldots$, 800m, we compute the translation and rotation errors per unit length of the trajectory. Translation error is reported as a percentage value, and rotation error in degrees per meter.

\vspace{.15cm}
\noindent
{\bf Kinect Data:} TUM Dataset~\cite{sturm12} consists of sequences of Kinect RGBD data captured in an indoor environment. The sensor resolution is 640x480, at 30 fps. The sequences come with a ground truth trajectory of the sensor, obtained from a high-accuracy motion-capture system. 
We test our algorithm on 7 sequences from the TUM dataset, down-sampled by a factor of 10 in both dimensions (i.e., 1/100\textsuperscript{th} the points). The error values for all three sequences are presented in Tab.~\ref{tab:TUM} (on page \pageref{tab:TUM}). We use two baseline methods: Super4PCS ~\cite{mellado14} + ICP and fast global registration (FGR)~\cite{zhou16} methods. Super4PCS works with point clouds without any point correspondences, and FGR uses point correspondences from the depth maps of the Kinect data. Both these methods are tested on full resolution Kinect data, whereas our approach only takes the sparsified input (down-sampled by a factor of 100). As shown in Tab.~\ref{tab:TUM}, we outperform the baselines in 6 out of the 7 sequences, despite using down-sampled data.

% \begin{figure*}[htb]
%     \centering
%     \includegraphics[width=0.5\linewidth]{figs/FullSeq3.png}%
%     \includegraphics[width=0.5\linewidth]{figs/FullSeq5.png}

%     \includegraphics[width=0.5\linewidth]{figs/FullSeq6.png}%
%     \includegraphics[width=0.5\linewidth]{figs/FullSeq7.png}
%     \caption{Visualization of the fully registered point-clouds from KITTI sequences 03, 05, 07, and 06 (clock-wise from top-left).}
%     \label{fig:full_reg_kitti}
% \end{figure*}
\vspace{0.15cm}
\begin{table*}[ht]
    \begin{center}
    \resizebox{.9\textwidth}{!}{
    \begin{tabular}{|p{0.25\linewidth}|c|c|c|c|c|c|}%
        \hline
        \multirow{2}{*}{\bf Sequence} & \multicolumn{2}{c|}{\bf Proposed} & \multicolumn{2}{c|}{\bf Super4PCS+ICP} &
        \multicolumn{2}{c|}{\bf FGR}
        \\ \cline{2-7}
%         & \multicolumn{2}{c|}{Proposed} & \multicolumn{2}{c|}{Super4PCS+ICP} &
%         \multicolumn{2}{c|}{FGR}
%         \\ \cline{2-7}
        & \pbox{0.11\linewidth}{tra. [mm]} & \pbox{0.11\linewidth}{rot. [\textdegree]} & \pbox{0.11\linewidth}{tra. [mm]} & \pbox{0.11\linewidth}{rot. [\textdegree]} & \pbox{0.11\linewidth}{tra. [mm]} & \pbox{0.11\linewidth}{rot. [\textdegree]} \\\hline \hline
        \texttt{fr1/xyz} & \textbf{3.86} & \textbf{0.46} & 12.99 & 0.49 & 9.51 & 0.70 \\
        \texttt{fr1/360} & 17.67 & 0.82 & 22.89 & 0.79 & {\bf 14.76} & {\bf 0.76} \\
        % \texttt{fr2/pioneer\_360} & 28.72 & {\bf 1.02} & {\bf 25.44} & 1.28 & 102.17 & 2.23 \\
        \texttt{fr3/sitting\_xyz} & \textbf{6.09} & 0.41 & 9.47 & \textbf{0.39} & 9.09 & 0.46 \\
        \texttt{fr1/room} & \textbf{7.25} & \textbf{0.52} & 15.29 & 0.73 & 11.82 & 0.70 \\
        \texttt{fr1/plant} & {\bf 6.47} & {\bf 0.61} & 14.23 & 0.74 & 11.86 & 0.68 \\
        \texttt{fr3/cabinet} & \textbf{7.25} & \textbf{0.55} & 16.56 & 0.64 & 16.63 & 0.84 \\
        \texttt{fr3/structure\-\_nn} & \textbf{4.49} & {\bf 0.35} & 8.28 & 0.53 & 9.77 & 0.67 \\ \hline
    \end{tabular}
    }
    \end{center}
\caption{\it Results on 7 sequences of the TUM RGBD dataset. Our method uses only the depth-maps, down-sampled by a factor of 100. Super4PCS + ICP and FGR use full resolution depth maps. Mean rotation and translation error between successive frames is reported.}
    \label{tab:TUM}
\end{table*}

\begin{figure}[ht]
    \centering
    \subfloat[Sequence 5]{\includegraphics[width=0.33\linewidth]{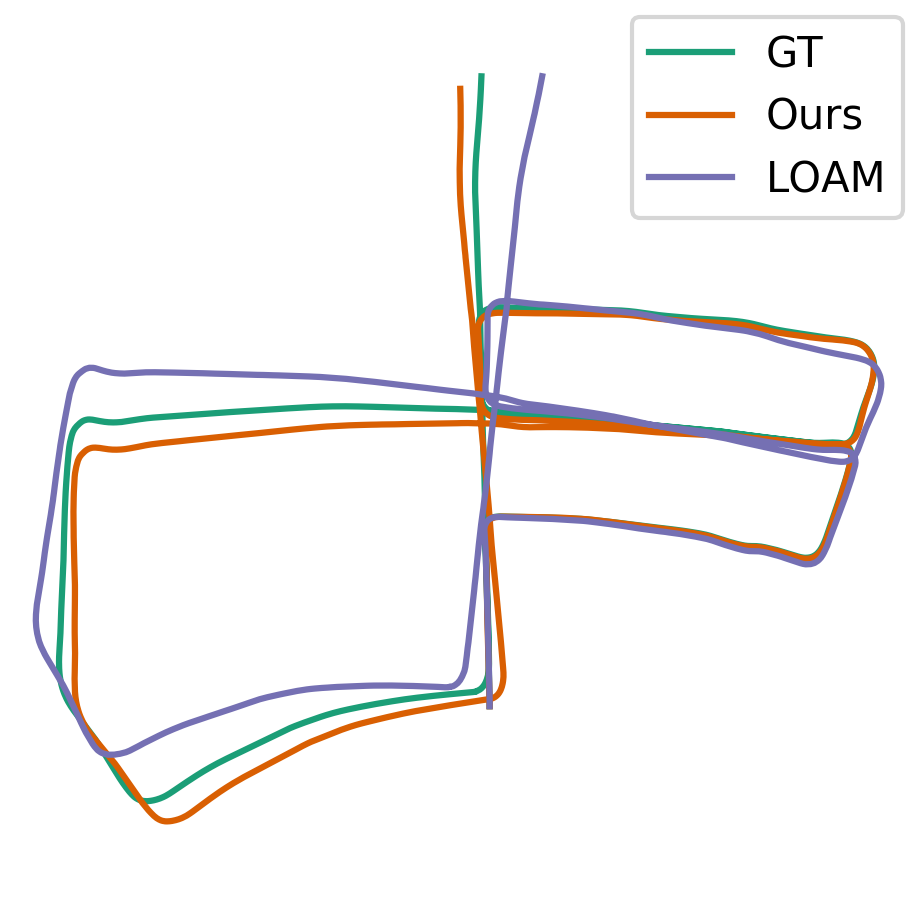}}
    \subfloat[Sequence 6]{\includegraphics[width=0.33\linewidth]{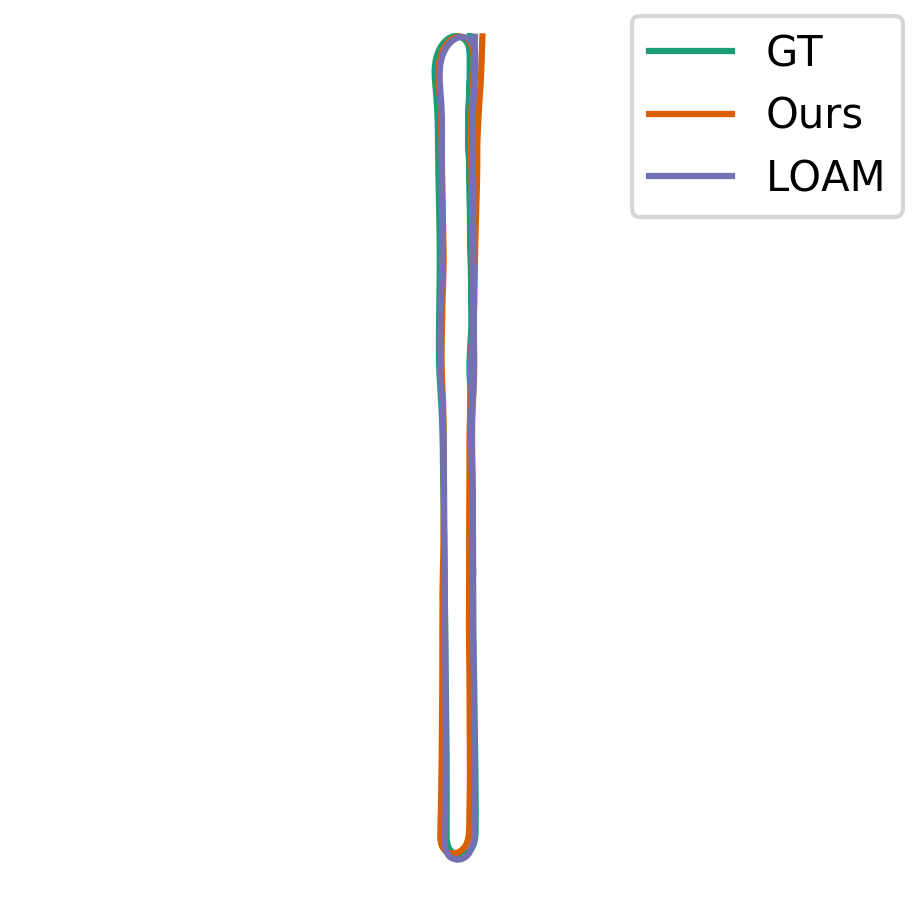}}
    \subfloat[Sequence 7]{\includegraphics[width=0.33\linewidth]{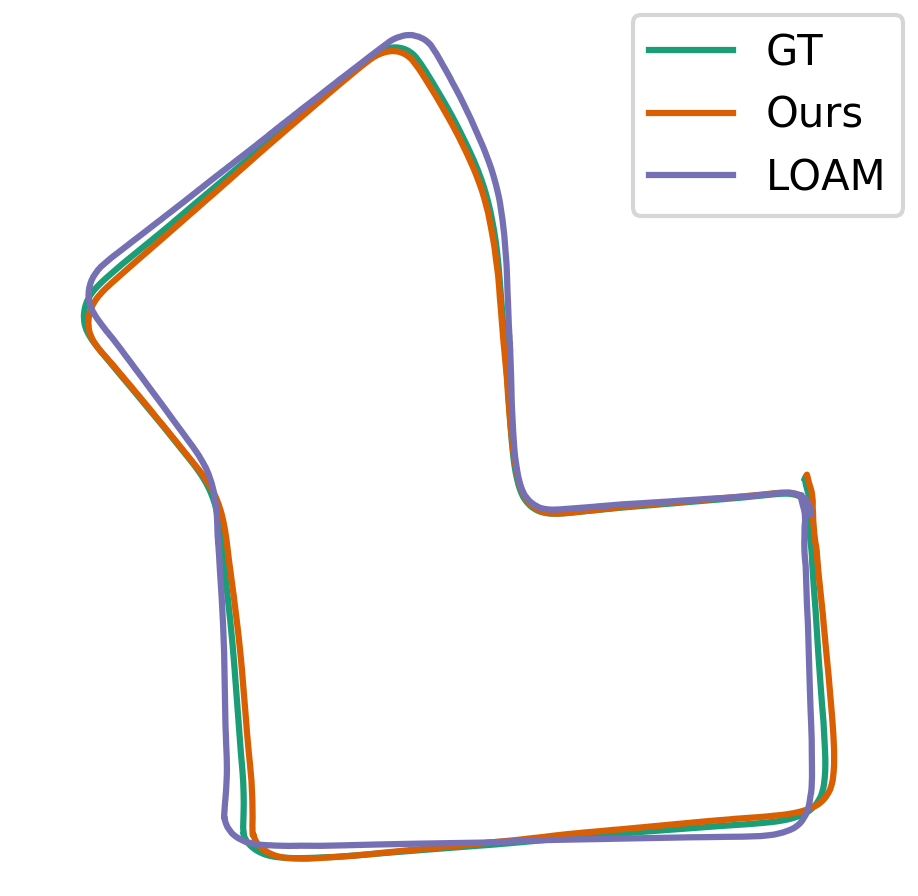}}
    \caption{\it Computed trajectories on KITTI Sequences at original resolution. The trajectory from the proposed method is closer to the ground truth than LOAM.}
    \label{fig:kitti_trajectories}
\end{figure}

\begin{table}[t]
\centering
\begin{tabular}{|c|c|c|c|c|c|c|c|c|}
\hline
\multirow{3}{*}{Seq.} & \multicolumn{4}{c|}{\bf Proposed}
& \multicolumn{4}{c|}{\bf LOAM} \\ \cline{2-9}
& \multicolumn{2}{c|}{Mean errors} & \multicolumn{2}{c|}{KITTI metrics}
& \multicolumn{2}{c|}{Mean errors} & \multicolumn{2}{c|}{KITTI metrics} \\ \cline{2-9}
& \multicolumn{1}{l|}{tra. [m]} & \multicolumn{1}{l|}{rot. [\textdegree]} & \multicolumn{1}{l|}{tra. [\%]} & \multicolumn{1}{l|}{rot. [\textdegree/m]} & \multicolumn{1}{l|}{tra. [m]} & \multicolumn{1}{l|}{rot. [\textdegree]} & \multicolumn{1}{l|}{tra. [\%]} & \multicolumn{1}{l|}{rot. [\textdegree/m]} \\
\hline \hline
00 & \textbf{0.024} & \textbf{0.105} & \textbf{1.612} & \textbf{0.007} & 0.061 & 0.457 & 2.865 & 0.015 \\
01 & 2.125 & 2.432 & 91.139 & 0.327 & \textbf{0.651} & \textbf{0.245}  & \textbf{25.711} & \textbf{0.022} \\
02 & 0.198 & \textbf{0.325} & 17.361 & 0.057 & \textbf{0.165} & 0.396 & \textbf{8.567} & \textbf{0.033} \\
03 & \textbf{0.035}  & \textbf{0.106} & \textbf{2.291} & 0.023 & 0.103 & 0.361 & 9.861 & \textbf{0.021} \\
04 & 0.698 & 0.247 & 57.950 & 0.091 & \textbf{0.138} & \textbf{0.213} & \textbf{1.927}  & \textbf{0.009} \\
05 & \textbf{0.018} & \textbf{0.096} & \textbf{1.407} & \textbf{0.009} & 0.048 & 0.312 & 1.721 & 0.011 \\
06 & \textbf{0.023} & \textbf{0.079} & \textbf{0.818} & \textbf{0.006} & 0.063 & 0.211 & 1.884 & 0.010 \\
07 & \textbf{0.018} & \textbf{0.084} & \textbf{0.999} & \textbf{0.006} & 0.042 & 0.321 & 2.080 & 0.014 \\
08 & \textbf{0.044} & \textbf{0.105} & 3.064 & \textbf{0.011} & 0.064 & 0.348 & \textbf{2.705} & 0.013 \\
09 & 0.249 & 0.366 & 18.654 & 0.054 & \textbf{0.109} & \textbf{0.328} & \textbf{5.196} & \textbf{0.031} \\
10 & \textbf{0.074} & \textbf{0.289} & 10.558 & 0.047 & 0.093 & 0.432 & \textbf{9.727} & \textbf{0.038} \\ \hline                  
\end{tabular}
\vspace{.15cm}
\caption{\it Mean translation and rotation errors between successive frames, and translation, rotation errors over the trajectory for KITTI sequences. Proposed method has a lower error than LOAM in many sequences.}
\label{tab:kitti_rpe}
\end{table}

\begin{table}[t]
\centering
\begin{tabular}{|c|c|c|c|c|c|c|c|c|} 
\hline
\multirow{3}{*}{Seq.} & \multicolumn{4}{c|}{\bf Proposed}
& \multicolumn{4}{c|}{\bf LOAM} \\ \cline{2-9}
& \multicolumn{2}{c|}{Mean errors} & \multicolumn{2}{c|}{KITTI metrics}
& \multicolumn{2}{c|}{Mean errors} & \multicolumn{2}{c|}{KITTI metrics} \\ \cline{2-9}
& \multicolumn{1}{l|}{tra. [m]} & \multicolumn{1}{l|}{rot. [\textdegree]} & \multicolumn{1}{l|}{tra. [\%]} & \multicolumn{1}{l|}{rot. [\textdegree/m]} & \multicolumn{1}{l|}{tra. [m]} & \multicolumn{1}{l|}{rot. [\textdegree]} & \multicolumn{1}{l|}{tra. [\%]} & \multicolumn{1}{l|}{rot. [\textdegree/m]} \\
\hline \hline
00 & \textbf{0.202} & \textbf{0.191} & \textbf{14.937} & \textbf{0.033} & 0.446 & 1.343 & 38.939 & 0.176 \\
01 & \textbf{2.196} & \textbf{0.936} & \textbf{94.428} & \textbf{0.134} & 2.205 & 1.602 & 95.274 & 0.212 \\
02 & \textbf{0.544} & \textbf{0.396} & \textbf{36.003} & \textbf{0.065} & 0.833 & 1.488 & 62.246 & 0.194 \\
03 & \textbf{0.277} & \textbf{0.300} & \textbf{33.545} & \textbf{0.064} & 0.583 & 1.573 & 89.949 & 0.491 \\
04 & \textbf{1.329} & \textbf{0.400} & \textbf{92.768} & \textbf{0.074} & 1.386 & 1.762 & 97.335 & 0.352 \\
05 & \textbf{0.172} & \textbf{0.180} & \textbf{14.139} & \textbf{0.038} & 0.527 & 2.041 & 54.738 & 0.280 \\
06 & \textbf{0.615} & \textbf{0.466} & \textbf{37.981} & \textbf{0.094} & 0.956 & 2.813 & 60.321 & 0.370 \\
07 & \textbf{0.123} & \textbf{0.123} & \textbf{14.522} & \textbf{0.038} & 0.334 & 1.491 & 36.841 & 0.209 \\
08 & \textbf{0.290} & \textbf{0.346} & \textbf{26.796} & \textbf{0.101} & 0.479 & 2.081 & 58.518 & 0.272 \\
09 & \textbf{0.551} & \textbf{0.541} & \textbf{44.765} & \textbf{0.134} & 0.781 & 1.921 & 66.405 & 0.255 \\
10 & \textbf{0.188} & \textbf{0.311} & \textbf{19.770} & \textbf{0.071} & 0.384 & 1.444 & 36.346 & 0.189 \\
\hline
\end{tabular}\vspace{.15cm}
\caption{{\bf Under extreme sparsity:} \it Mean translation and rotation errors between successive frames, and translation, rotation errors over the trajectory for KITTI sequences at low resolution ($1/36^{\text{th}}$ the points). Proposed method outperforms LOAM in all sequences.}
\label{tab:kitti_rpe_d36}
\end{table}

\begin{figure*}[ht]
    \centering
    \includegraphics[width=.9\linewidth]{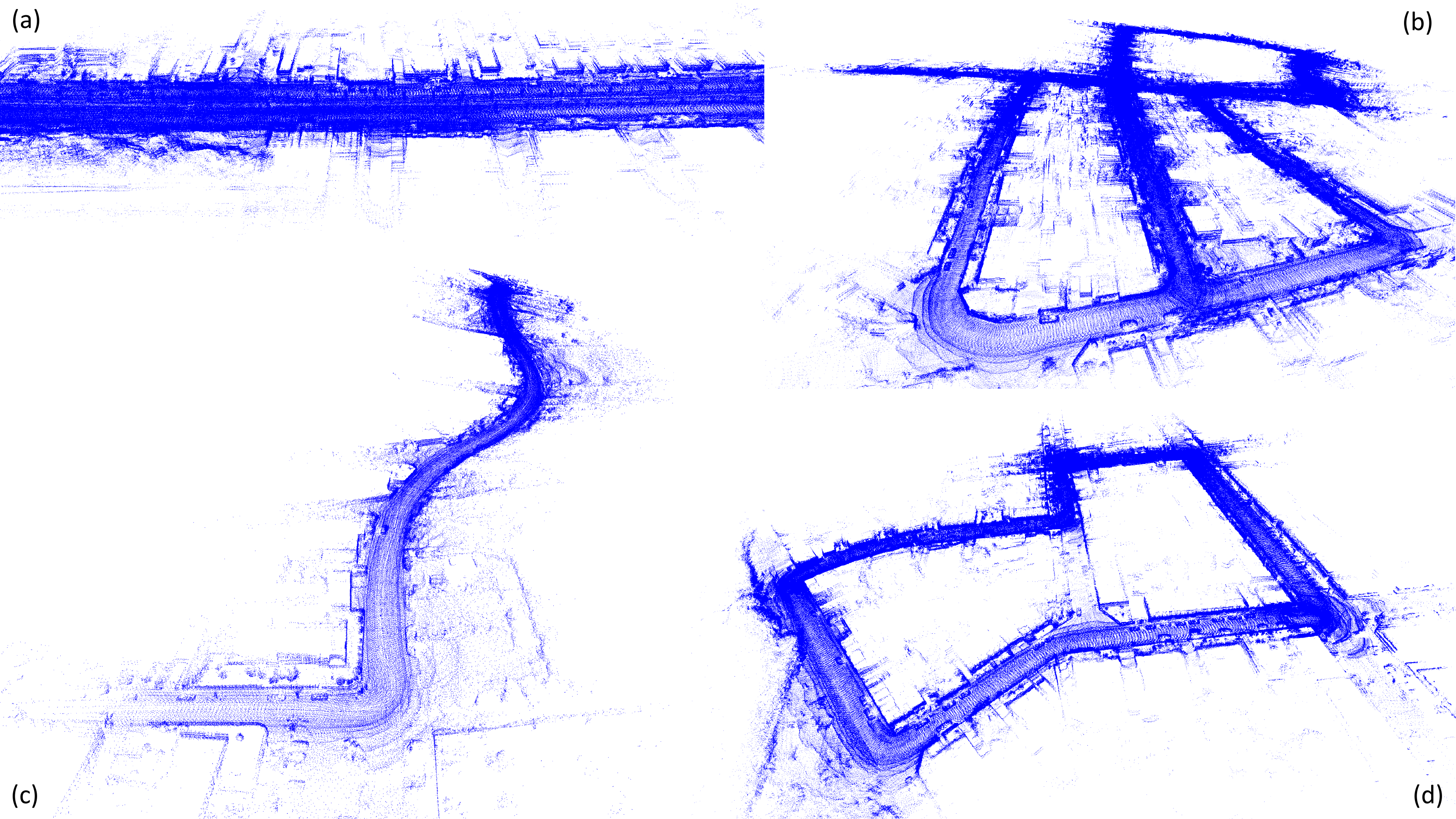}%
    \caption{\it Visualization of the fully registered point-clouds from KITTI sequences (a) 06, (b) 05, (c) 03, and (d) 07.}
    \label{fig:full_reg_kitti}
\end{figure*}

\vspace{.15cm}
\noindent
{\bf LiDAR Data:} KITTI dataset~\cite{Geiger2012CVPR} consists of LiDAR point-clouds collected from the top of a moving vehicle. The LiDAR sensor captures roughly 10 fps (frames per second), with about 100k points per frame.

We compare our results against the LOAM Algorithm~\cite{Zhang14}, which is currently highly ranked on the online leaderboard. The computed trajectories on these sequences are shown in Fig.~\ref{fig:kitti_trajectories}, the relative pose error between consecutive frames, and the error along the trajectory is reported in Tab.~\ref{tab:kitti_rpe} (on page~\pageref{tab:kitti_rpe}). % Trajectory error is measured as a fraction of the distance traveled on trajectory segments of lengths 100m, 200m, \ldots, 800m, as specified in \cite{Geiger2012CVPR}.
Selected full point-clouds after registration are presented in Fig.~\ref{fig:full_reg_kitti} (on page \pageref{fig:full_reg_kitti}).
%{\color{blue}PM: Can we visualize differences in the registration using our method and LOAN? If yes, we could add that in Fig. 7.}.

The results of %\textcolor{red}{the LOAM algorithm} \textcolor{blue}{LOAM}
LOAM here differ from those on the leaderboard because we run LOAM by ourselves, without using IMU data, for a fair comparison. 
%\textcolor{red}{We use the open-source version of LOAM available at \url{https://github.com/laboshinl/loam_velodyne}, the official version is no longer available.} \textcolor{blue}{
As the official version is unavailable, we use the open-source version of LOAM \footnote{ \url{https://github.com/laboshinl/loam_velodyne}}.
%}

\vspace{.15cm}
\noindent
\setlength{\intextsep}{0pt}%
\setlength{\columnsep}{7pt}%
\begin{wraptable}{r}{0.5\textwidth}
% \begin{table}
\centering
\resizebox{.45\textwidth}{!}{
\begin{tabular}{|l|c|c|}
\hline
{\bf Algorithm}    & tra. err. [\%]  & rot. err. [deg/m]  \\
\hline \hline
7L           & 2.628          & 0.021             \\
5L1C         & 2.452          & \textbf{0.017}    \\
3L2C         & 2.589          & 0.025             \\
1L3C         & 2.794          & 0.021             \\
\hline
Combined     & \textbf{2.291} & 0.023             \\ \hline
\end{tabular}}
\caption{\it Comparison of our algorithms}
\label{tab:kitti_comparison}
% \end{table}
\end{wraptable}
{\bf Comparison of Our Algorithms:}
We test all our algorithms separately on sequence 03 of the KITTI dataset. The combined approach produces the best translation and 5L1C produces the best rotation (Tab.~\ref{tab:kitti_comparison}). All of these approaches have their own strengths and weaknesses. For example, line intersection constraints handle planar regions better, and the methods utilizing corner points are supposed to handle scenes containing vegetation and other non-planar regions.

\setlength{\intextsep}{0pt}%
\setlength{\columnsep}{7pt}%
\begin{wrapfigure}{r}{0.35\textwidth}
% \begin{figure}[ht]
\vspace{-.65cm}
    \centering\captionsetup{justification=centering}
    \includegraphics[width=0.7\linewidth]{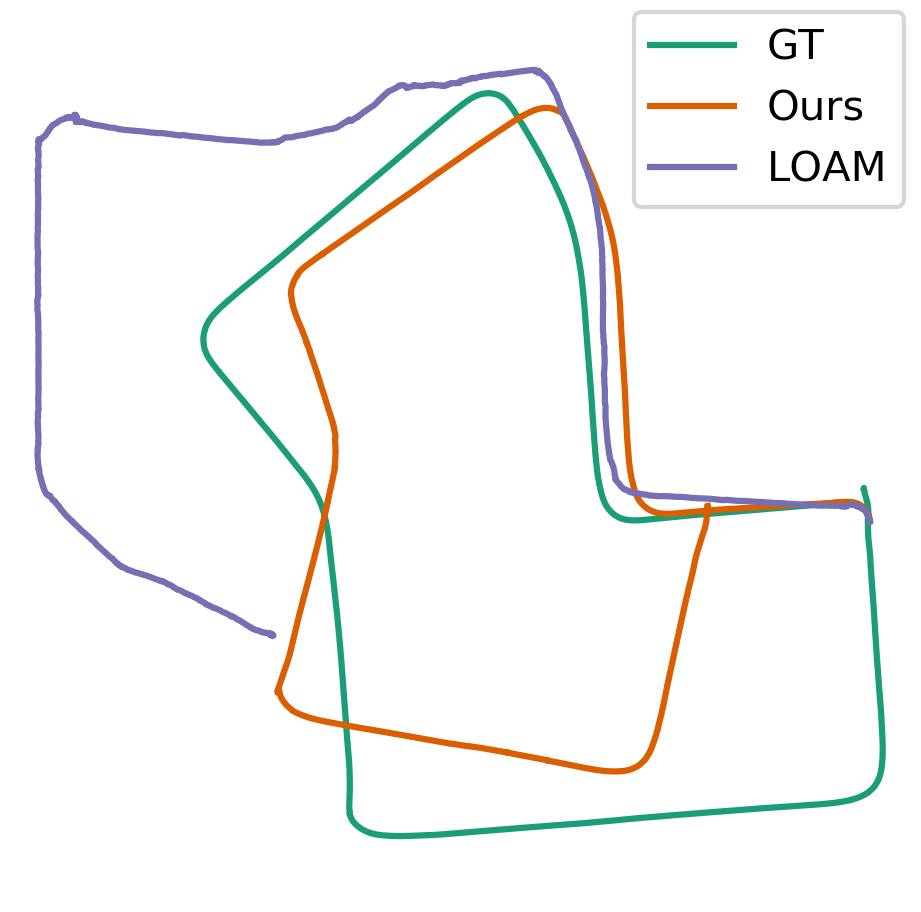}
    \caption{\it 1/36 Down-sampled KITTI sequence 7}
    \label{fig:kitti_07_down}
% \end{figure}
\end{wrapfigure}
\vspace{.15cm}
\noindent
{\bf Under extreme sparsity:} We test our approach on LiDAR data down-sampled by a factor of 6 in both dimensions, $1/36^{\text{th}}$ of all points  (see Fig.~\ref{fig:downsampling}, page~\pageref{fig:downsampling}) and compare our results with LOAM on the KITTI dataset. We show a trajectory in Fig.~\ref{fig:kitti_07_down}, and the errors are reported in Tab.~\ref{tab:kitti_rpe_d36} (on page \pageref{tab:kitti_rpe_d36}). While both our algorithm and LOAM deteriorate as we increase the sparsity, our algorithm significantly outperforms LOAM. %As shown in Fig.~\ref{fig:kitti_07_down}, our approach estimates a reasonable trajectory with a very small rotation error (0.1deg/m) even on 64x down-sampled data.

% \begin{table}[htb]
%     \centering
%     % \resizebox{.98\textwidth}{!}{
%     \begin{tabular}{|p{2.25cm}|c|c|c|c|}
%         \hline
%         \multirow{2}{\linewidth}{\parbox{2.25cm}{\centering Down-sampling factor}} & \multicolumn{2}{c|}{\bf Proposed} & \multicolumn{2}{c|}{\bf LOAM} \\ \cline{2-5}
%         & \multicolumn{1}{p{0.1\linewidth}|}{tra. [\%]} & \multicolumn{1}{p{0.145\linewidth}|}{rot. [deg/m]} & \multicolumn{1}{p{0.1\linewidth}|}{tra. [\%]} & \multicolumn{1}{p{0.145\linewidth}|}{rot. [deg/m]} \\ \hline \hline
%         36 & \textbf{7.419} &\textbf{0.023} & 52.232 & 0.349 \\
%         64 & \textbf{23.166} & \textbf{0.107} & 58.681 & 0.494 \\ 
%         \hline
%     \end{tabular}
%     % }
%     \caption{Results on down-sampled KITTI sequence 07}
%     \label{tab:kitti_07_down}
% \end{table}

\vspace{.15cm}
\noindent
{\bf Performance and Speed:} Our code is implemented in C++. The average computation time 
%\textcolor{blue}{
on KITTI dataset at full resolution
%} 
for various operations (running on 1 core of Intel i7-8700K) is as follows: 22ms for the {\sc AP} solver; 2ms for the inlier counting; and 3s for the line fitting. The time taken by inlier counting varies with the number of detected lines; these are typically observed values. 
%\textcolor{blue}{
The total runtime of the entire algorithm depends on the number of RANSAC iterations and can be brought down to 30s by paralleling on 10 threads.
%}% (our method is highly parallelizable). %We also explored an early stop strategy which gets us a 1.5x speedup with a slight performance hit. 

% given in table~\ref{tab:speed}. 

% \begin{table}[htb]
%     \centering
%     \scalebox{.8}{
%     \begin{tabular}{P{.16\linewidth}|P{.16\linewidth}|P{.16\linewidth}|P{.16\linewidth}|P{.16\linewidth}|P{.16\linewidth}}
%          & 7 Lines & 3 Lines \& 1 Plane & 1 Line \& 2 Plane & Inlier counting & line fitting \\ \hline
%         Time [$\mu s$] & 22k (=0.022 s) &  4.9 & 0.64 & 10 to 2000 & 3M (=3s) \\
%     \end{tabular}}
%     \caption{Average computation time for each of the proposed solvers, and other tasks.}
%     \label{tab:speed}
% \end{table}

We use a threshold (2cm on KITTI, 5mm on TUM) as well as max number of iterations (30K iterations) for terminating the {\sc AP}. In the future, we will explore certain extrapolated projection techniques for further speedup~\cite{Censor2012}.

\subsection{Sensitivity analysis}
We conducted some experiments to study the robustness of our algorithm to large motion and sparse settings.

\setlength{\intextsep}{0pt}%
\setlength{\columnsep}{7pt}%
\begin{wraptable}{r}{0.5\textwidth}
% \begin{table}
\centering
\resizebox{.45\textwidth}{!}{
\begin{tabular}{|c|c|c|c|}
\hline
tra. [m] & succ. rate  & rot. [deg]  & succ. rate \\
\hline \hline
0.5 & 94\%        & 0.5     & 97\%            \\
2   & 86\%        & 2       & 91\%            \\
8  & 79\%         & 8       & 83\%             \\
32 & 65\%         & 32      & 78\%             \\
128  & 60\%       & 128     & 49\%             \\ \hline
\end{tabular}}
\caption{\it Robustness to large translations and rotation by augmenting them in noisy LiDAR data. }
\label{tab:convergence}
% \end{table}
\end{wraptable}
\vspace{.15cm}
\noindent
{\bf Convergence under large motion: } 
We generate a pair of frames by taking the first frame of KITTI's sequence 7 and apply large translation and rotation to it. Then we use the same fitting strategy to get pairs of lines and edge/corner candidates from the two frames. We repeat the process 100 times and applied the AP solver for registration. We count a success when the registration has rotation error smaller than 0.5\textdegree and translation error within 0.1 meter. The success rate is the number of success out out of 100 trials. As shown in Tab.~\ref{tab:convergence}, the success rate is monotonic and RANSAC only needs some good hypotheses. For comparison, the maximum translation, rotation between two frames in KITTI is 1.5 m and 5\textdegree respectively. % The only catch under large transformation is the correspondences in texture-less and sparse settings, and again RANSAC can deal with a significant number of outliers. 

\begin{wrapfigure}{r}{0.4\textwidth}
% \begin{figure}[!htbp]
   \includegraphics[height=3.0 cm]{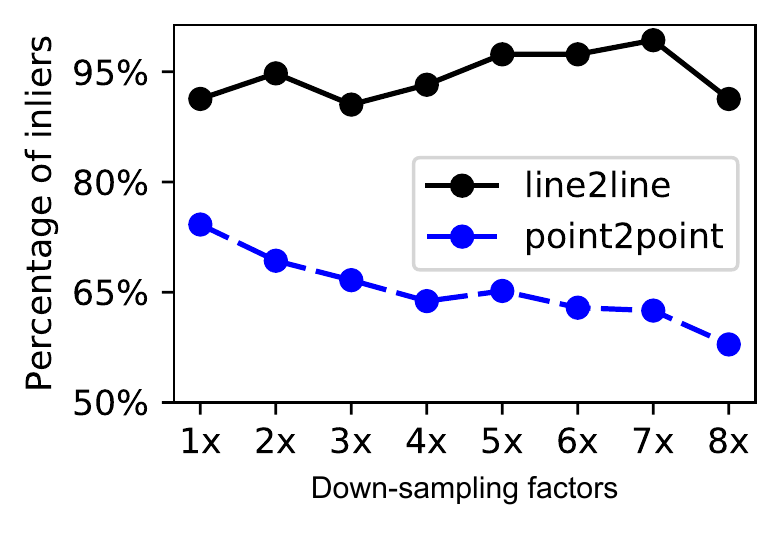}
   \caption{\it Inlier ratio of point correspondence vs. intersected lines as sparsity increases.}
   \label{fig.inlier_ratio}
% \end{figure}
\end{wrapfigure}
\vspace{.15cm}
\noindent
{\bf Robustness of intersection constraints under sparsity:}
Under sparsity, line intersection constraints continue to exist while point correspondences diminish significantly. We take two consecutive frames of KITTI's sequence 7 and align them using the ground truth.
% We show that the line intersection constraints fit the sparse point-cloud registration better than point correspondence. We take two point-clouds/frames from KITTI. 
% This dataset contains rich urban scenes which are predominantly planar, enabling a large mount of intersecting line segments. 
For each line/point in the first frame, we find the closest line/point in the second frame. We fixed a threshold for the line/point distance for inlier counting and vary the down-sampling factor (sparsity). As observed in Fig.~\ref{fig.inlier_ratio}, the percentage of inliers using line intersections (line2line in Fig.~\ref{fig.inlier_ratio}) is not affected by sparsity. On the other hand, the percentage of inliers using point correspondences (point2point in Fig.~\ref{fig.inlier_ratio}) decreases considerably. This means that point-based methods like ICP are more sensitive to the inlier threshold.
       % results
\section{Discussion}\label{sec:discussion}

The proposed algorithm applies to sensors on a moving platform and we make smoothness assumption for obtaining line intersection constraints, although our algorithm is robust to outliers due to the use of near-minimal solvers in a RANSAC framework. While we outperform LOAM in 6 out of 11 sequences, our method can further be improved by following LOAM and correcting for distortions obtained from moving platforms~\cite{Zhang14}. We also show that we outperform LOAM in all the KITTI sequences under extreme sparsity. Note that the proposed method is not customized to a specific sensor. After handling the challenges in LiDAR, we ported to Kinect datasets with almost zero-development cost and outperformed all baselines in 6 out of 7 sequences using 1/100th of the data, without texture. 

At the heart of our technique, we hinge on two elements to achieve the superior performance. First, we exploit interior regional information (by fitting line segments on the point-cloud and generating rich line intersection constraints) in addition to the boundary information (through the extraction of corners and edges) typically used by classical methods. Every line intersection adds a planarity constraint implicitly on the four end points. Compared to plane detection, intersected lines is easier to find in sparse point cloud. Plane detection in sparse settings with sensor motion distortion can be brittle, and finding enough planes in near-degenerate situations will be challenging (e.g., many road scenes with buildings on both sides or highways may not provide enough planes to lock the pose from sliding). Second, despite the simplicity, {\sc AP} should not be treated as a trivial endeavour. In addition to several vision related problems~\cite{Zhou2015a,Zhou2015b,yan15,Schops2019,Campos2019}, variants of {\sc AP} have also been used to solve computationally hard problems such protein folding, sphere packing, and Sudoku~\cite{Gravel_2008,Elser_2007}.

In our work, we observed that {\sc AP} can solve near-minimal problems that are known to be hard for algebraic solvers. From the formulations used in this paper, it is not difficult to see that the proposed approach can be easily extended to other geometric vision problems as well. We show a video demonstration of our algorithm in the Supplementary Materials.    % conclusions

% acknowledgments
\section*{Acknowledgments}
We thank the anonymous reviewers for valuable feedback. This work was partially supported by the National Science Foundation (NSF) grant IIS 1764071; the Portuguese National Funding Agency for Science, Research and Technology LARSyS - FCT Plurianual funding 2020-2023; and the Polish operational program IDOP, project ``Development of an innovative, autonomous mobile robot for retail stores''.

\bibliographystyle{splncs}
\bibliography{main}

\end{document}